%% file: main.tex
\documentclass[runningheads]{llncs}

\usepackage{eccv}

\usepackage{eccvabbrv}

\usepackage{graphicx}
\usepackage{booktabs}

\usepackage{lipsum}
\usepackage{colortbl}
\usepackage{makecell}
\usepackage{multirow}

\usepackage{pifont}
\usepackage{bbding}  

\usepackage{amsmath}  
\usepackage{amssymb}  

\usepackage{xcolor}
\definecolor{V}{RGB}{21,137,139}
\definecolor{X}{RGB}{234,120,60}

\usepackage[linesnumbered,ruled,vlined]{algorithm2e}

\usepackage[pagebackref,breaklinks,colorlinks,citecolor=eccvblue]{hyperref}

\usepackage{orcidlink}

\begin{document}

\title{FVG-PT: Adaptive Foreground View-Guided Prompt Tuning for Vision-Language Models} 

\titlerunning{Adaptive FVG-PT for Vision-Language Models}

\author{Haoyang Li\inst{1,2}\orcidlink{0000-0002-8344-1201} \and
Liang Wang\inst{1,2} \and
Siyu Zhou\inst{1} \and
Jiacheng Sun\inst{2} \and
Jing Jiang\inst{1} \and
Chao Wang\inst{2}\thanks{Corresponding authors.} \and
Guodong Long\inst{1}\protect\footnotemark[1]\orcidlink{0000-0003-3740-9515} \and
Yan Peng\inst{2}\protect\footnotemark[1]\orcidlink{0000-0003-1312-9527}
}

\authorrunning{H.~Li et al.}

\institute{
University of Technology Sydney, NSW 2007, Australia \and
Shanghai University, Shanghai 200444, China \\
\email{haoyang.li-3@student.uts.edu.au}
}
\maketitle

\begin{abstract}
  CLIP-based prompt tuning enables pretrained Vision-Language Models (VLMs) to efficiently adapt to downstream tasks. Although existing studies have made significant progress, they pay limited attention to changes in the internal attention representations of VLMs during the tuning process. In this paper, we attribute the failure modes of prompt tuning predictions to shifts in foreground attention of the visual encoder, and propose \textbf{F}oreground \textbf{V}iew-\textbf{G}uided \textbf{P}rompt \textbf{T}uning (\texttt{FVG-PT}), an adaptive plug-and-play foreground attention guidance module, to alleviate the shifts. Concretely, \texttt{FVG-PT} introduces a learnable Foreground Reliability Gate to automatically enhance the foreground view quality, applies a Foreground Distillation Compensation module to guide visual attention toward the foreground, and further introduces a Prior Calibration module to mitigate generalization degradation caused by excessive focus on the foreground. Experiments on multiple backbone models and datasets show the effectiveness and compatibility of \texttt{FVG-PT}. Codes are available at: \href{https://github.com/JREion/FVG-PT}{https://github.com/JREion/FVG-PT}.
\end{abstract}

\section{Introduction}
\label{sec:intro}

In recent years, Vision-Language Models (VLMs) such as CLIP \cite{radford2021clip} have achieved strong cross-modal alignment and zero-shot generalization on open-set recognition and multi-modal perception tasks. To adapt pretrained CLIP to specific classification-related downstream tasks, prompt tuning is proposed as a parameter efficient strategy \cite{han2024peft, xing2024peft2}. This approach keeps all parameters of the foundation CLIP frozen and introduces lightweight learnable prompts as input queries, thereby guiding the model outputs toward the target task distribution \cite{zhou2022coop}.

Recent studies propose various forms of prompt tuning, including text prompts \cite{zhou2022cocoop, li2025atprompt}, visual prompts \cite{jia2022vpt, pei2024sa2vp}, cross-modal prompts \cite{khattak2023promptsrc, li2025dpc}, and mid-layer plugins of encoders \cite{seputis2024mma, guo2025mmrl}. Despite their success, these methods typically focus on refining the design of learnable prompts or the interaction patterns between visual and textual features, while paying limited attention on the impact of prompt tuning process on the internal attention representations of VLMs.

\begin{figure}[t]
  \centering
  \includegraphics[width=0.95\textwidth]{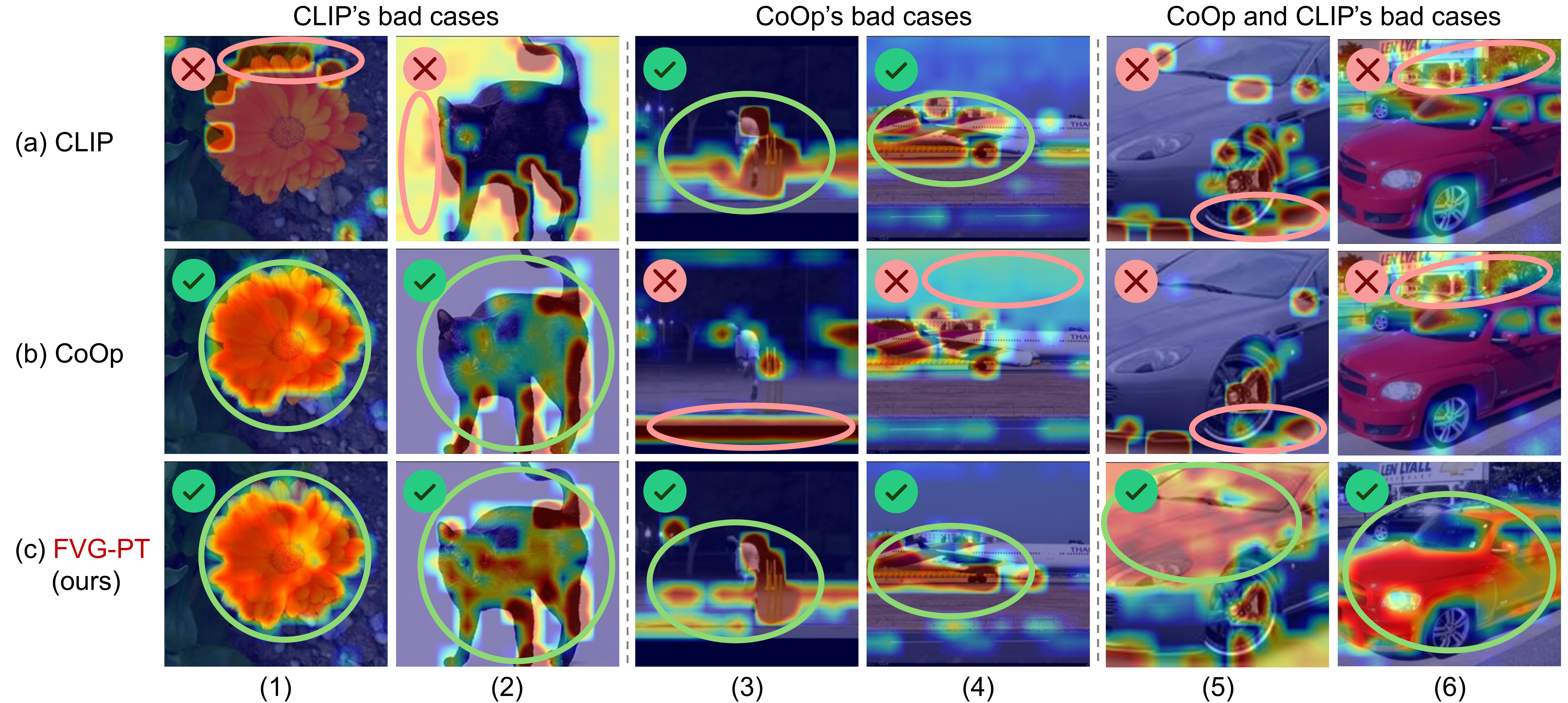}
  \caption{Comparison of visual encoder attention maps generated by Grad-CAM \cite{selvaraju2017gradcam} for the same image under (a) original CLIP \cite{radford2021clip}, (b) CoOp \cite{zhou2022coop}, and (c) our \texttt{FVG-PT}. In the bad cases of (a) and (b), attention deviates from the foreground view, where \texttt{FVG-PT} effectively suppresses this shift and leads to a correct prediction.}
  \label{Fig-1}
\end{figure}

Specifically, because CLIP performs image-text alignment through contrastive learning \cite{radford2021clip}, the text prompt used as a query determines the response weights of image patch tokens in the feature space, which implies that the input prompts directly shape the internal attention distribution over the image (details are in Sec. \ref{sec2}). Therefore, during prompt tuning, as the learnable prompts are continuously optimized, the corresponding image attention distribution changes accordingly. To study how this evolution influences the predictions, we conduct Grad-CAM-based \cite{selvaraju2017gradcam} case studies and analyze the attention behavior of prompt tuning method (CoOp \cite{zhou2022coop}) before and after fine-tuning. We focus on samples where CLIP and CoOp produce inconsistent predictions to characterize failure cases of prompt tuning. As in Fig.~\ref{Fig-1}(a-b), we observe that for misclassified images, the attention in the visual ViT encoder often deviates from the main object (\textbf{foreground}) to irrelevant background, thereby preventing the foreground semantics from aligning well with text. Quantitative analysis in \textit{Supplementary Material~\ref{sec:B4}} further shows that higher-performing models tend to exhibit smaller foreground shifts. These results suggest that \textbf{many incorrect predictions in prompt tuning can be attributed to shifted allocation of foreground attention of the visual encoder}. As a result, to mitigate such shift, it is crucial to guide attention to focus on the image foreground during prompt tuning.

Although some researches make initial attempts to guide VLM attention toward the foreground (e.g., by leveraging gradients \cite{zhu2023prograd} or explicit visual supervision \cite{shtedritski2023redcircle, yang2023fgvp, zhang2025dapt}), several key limitations remain unaddressed: (1) the quality of the foreground view is not controllable and may be incorrect, incomplete, or overly large, which can lead the model to learn inaccurate alignment patterns; (2) since image backgrounds often contain commonsense features \cite{an2023backgroundCLIP, zhang2025dapt}, over-focusing attention on the foreground in target (\textbf{base}) classes may weaken these commonsense, thereby harming generalization to unseen (\textbf{new}) classes, which is referred to as the base-new trade-off (BNT) \cite{zhang2024dept, li2025dpc, li2025mao}. In summary, the central bottleneck of foreground-guided methods is the lack of an \textbf{adaptive mechanism}: the model cannot flexibly adjust its trust in the foreground view or dynamically balance base-class adaptation and new-class generalization.

To address these issues, we propose \textbf{F}oreground \textbf{V}iew-\textbf{G}uided \textbf{P}rompt \textbf{T}uning (\texttt{FVG-PT}), a plug-and-play prompt tuning framework that adaptively controls the quality of foreground view, balances base and new branches, and integrates seamlessly with diverse prompt tuning backbones. To obtain high-quality foreground views, we first apply a pretrained external segmentation model \cite{zou2023seem} and extract explicit foreground regions of the input image from semantic masks. On top of this prior, we introduce a learnable \textbf{\textit{Foreground Reliability Gate}}, implemented as a multi-layer perceptron (MLP) that takes multiple foreground reliability indicators as input and outputs a scalar coefficient $r$ as the foreground trust score. This gate encourages the model to learn an adaptive weighting scheme for foreground features under a set of combined criteria. Since the gate is class-agnostic, it can assess the reliability of any foreground view input.

To guide VLM attention toward the foreground, we then design a \textbf{\textit{Foreground Distillation Compensation}} module driven by $r$. We insert lightweight learnable adapters \cite{houlsby2019Adapter} into the visual and textual branches of frozen pretrained backbone models (e.g., CoOp \cite{zhou2022coop}) after original feature alignment stage, with the goal of learning a foreground-oriented feature re-projection pattern on top of the backbones, while remaining compatible with the backbones' original optimization direction. Finally, to overcome the BNT problem, we propose a \textbf{\textit{Prior Calibration}} module that structurally decouples the new branch from foreground-enhanced base branch, thereby isolating their optimization paths. In the new branch, we introduce Backbone Reliability Gate that learns adaptive weights between the base-branch backbone and the original CLIP prior, preserving sufficient commonsense and improving generalization to new classes.

In experiments, we attach \texttt{FVG-PT} to 4 backbone models \cite{zhou2022coop, yao2023kgcoop, khattak2023promptsrc, guo2025mmrl} and evaluate it on 11 datasets, which verifies the effectiveness of optimizing prompt tuning through foreground view and its compatibility. The main contributions of this paper are summarized as follows.

\begin{itemize}
    \item We conduct an insightful analysis that attributes the failure modes of prompt tuning to the attention shifts of the visual encoder relative to the foreground, and propose Foreground View-Guided Prompt Tuning to introduce explicit foreground supervision to alleviate this shift, thereby achieving plug-and-play prompt tuning enhancement.
    
    \item To guide visual attention to correct foreground regions, we propose a Foreground Reliability Gate that adaptively addresses uncertainty in the foreground view, and design a Foreground Distillation Compensation module to provide plug-and-play attention guidance.
    
    \item We propose Prior Calibration module, which decouples base and new branches and adaptively balances the backbone and CLIP prior, mitigating the negative effect of foreground-focused learning on new-class generalization.
\end{itemize}

\section{Related Work}  \label{sec2}

\paragraph{\bf Prompt Tuning in VLMs.}  Vision-Language Models (VLMs) \cite{lu2019vilbert, radford2021clip, jia2021align, li2023blip, li2026hardnegative} receive significant attention for their ability to deeply align and fuse image and text representations, achieving strong performance on a wide range of cross-modal perception and understanding tasks. Given the high computational cost of training and fine-tuning VLMs due to their large parameter scale and data requirements, prompt tuning is proposed as a parameter-efficient alternative that adapts pretrained VLMs \cite{han2024peft}. Unlike hand-crafted hard prompt templates \cite{li2021prefix, wei2022chain} in LLMs or CLIP (e.g., ``\textit{A photo of a [CLASS]}''), prompt tuning introduces a set of lightweight learnable vectors as queries to the frozen model, guiding its outputs toward domain-specific data through continuous optimization. Existing work primarily focuses on improving the design of learnable prompts to enhance task-specific image-text alignment, including text prompts \cite{zhou2022coop, tian2024argue}, visual prompts \cite{jia2022vpt, pei2024sa2vp, xu2024provp}, cross-modal prompts \cite{zang2022upt, khattak2023promptsrc, li2024promptkd, li2025augpt}, and mid-layer plugins of encoders \cite{seputis2024mma, guo2025mmrl}. More recent studies incorporate external knowledge into prompt tuning \cite{yao2023kgcoop, roy2023coprompt, li2025atprompt}. In contrast, our work focuses on guiding the visual encoder's attention, offering a plug-and-play solution that is orthogonal to these approaches.

\paragraph{\bf Visual Attention Guidance for CLIP.} In prompt tuning, the changes in visual attention representations deserve closer examination. Specifically, in CLIP, the image and text inputs are encoded into global features using Transformer-based \cite{dosovitskiy2020vit} encoders, and classification is performed based on image-text similarity. The visual encoder constructs global image representations by selectively aggregating patch features through self-attention, emphasizing regions that exhibit higher similarity to the corresponding text features \cite{radford2021clip}. Under this alignment paradigm, the prompt (whether textual or visual) serves as a semantic query that shapes the model's similarity-driven preferences over different image regions. Therefore, variations in the prompt alter the similarity objective, leading to a reallocation of response weights across image patches and resulting in observable shifts in the visual attention distribution. As prompt tuning involves continuously optimizing learnable prompts, tracking and analyzing these attention dynamics is particularly important.

As shown in Fig.~\ref{Fig-1}, our empirical analysis confirms that guiding visual attention toward the image foreground can enhance prompt tuning. Existing methods attempt to influence the visual attention of foundation CLIP model by optimizing gradient propagation \cite{zhu2023prograd} or adding visual cues (e.g., visual markers \cite{shtedritski2023redcircle, zhuang2024falip}, alpha channels \cite{sun2024AlphaCLIP}, or foreground boundaries \cite{yang2023fgvp, zhang2025dapt}). However, most of these approaches do not provide explicit evaluation or control over the guidance process (e.g., assessing the quality of the foreground view). In this paper, we address this gap by introducing an adaptive Foreground Reliability Gate, followed by a Foreground Distillation Compensation module to enable plug-and-play foreground attention guidance across multiple prompt tuning backbones.

\paragraph{\bf Base-New Trade-off.} In prompt tuning, the base-new trade-off (BNT) problem receives considerable attention \cite{zhou2022cocoop}. It can be attributed to the overfitting of learnable prompts to the target task (base classes) during optimization, which leads to reduced generalization on new classes. In this paper, BNT manifests as excessive attention to the foreground, which may cause the model to overlook commonsense knowledge embedded in the background. Existing approaches attempt to mitigate BNT through conditional context \cite{zhou2022cocoop, yao2024tcp}, consistency constraints \cite{khattak2023promptsrc}, external knowledge \cite{khattak2024protext, wang2024hpt, li2024promptkd}, or decoupling at the feature or prompt level \cite{zhang2024dept, li2025dpc, zhang2025dapt}. In contrast, our \texttt{FVG-PT} introduces a model-agnostic Prior Calibration strategy that fully decouples the foreground (base) and new branches, and learns an adaptive balance between the backbone model and the CLIP prior, resulting in consistent improvements on both base and new classes.

\begin{figure}[t]
  \centering
  \includegraphics[width=\textwidth]{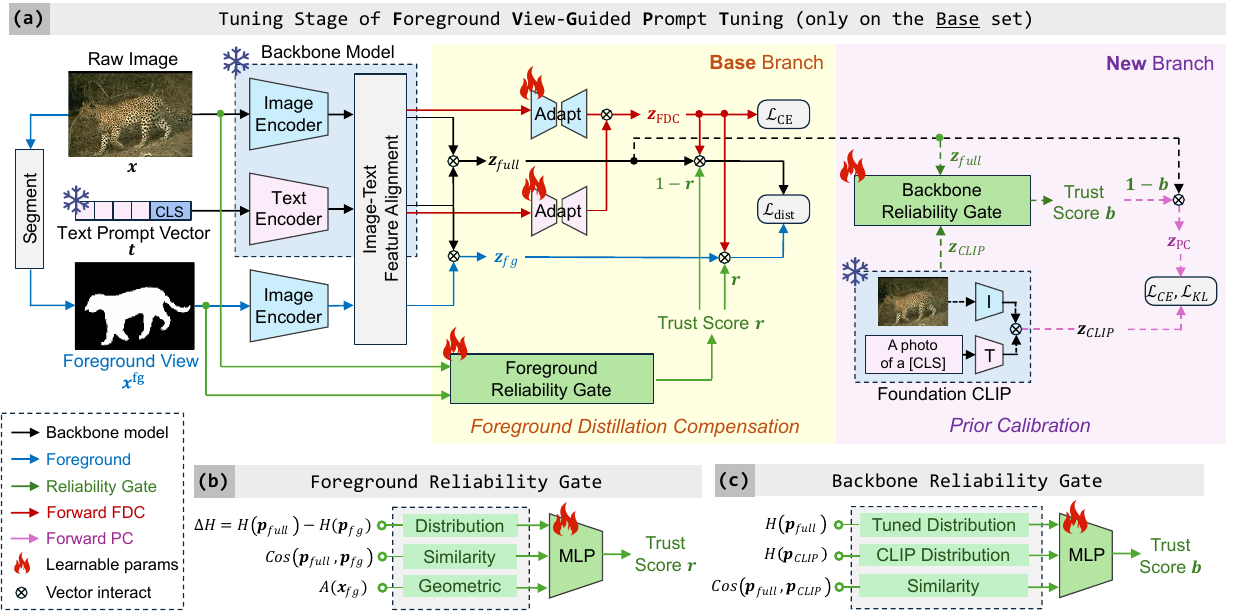}
  \caption{Framework of our proposed \texttt{FVG-PT}. As a plug-and-play method, in (a) tuning stage, \texttt{FVG-PT} obtains the foreground view $x^{\mathrm{fg}}$ of image $x$ and fine-tunes the (b) Foreground Reliability Gate to learn a foreground trust score $r$. Meanwhile, Foreground Distillation Compensation module inserts adapters after image-text alignment of frozen backbone model to guide visual attention toward the foreground. In parallel, independent Prior Calibration fine-tunes the (c) Backbone Reliability Gate on new branch (indicated by \textit{dashed lines}) to balance the tuned model and the CLIP prior.}
  \label{Fig-2}
\end{figure}

\section{Proposed Method}  \label{sec3}

The framework of \texttt{FVG-PT} is illustrated in Fig.~\ref{Fig-2}. As a plug-and-play enhancement method, our approach is built upon pre-tuned backbone models (e.g., CoOp \cite{zhou2022coop}). For an input image-text pair $(x, t)$ used in backbone, \texttt{FVG-PT} first obtains the foreground view $x^{\mathrm{fg}}$ of the image $x$. Next, guided by the adaptive trust score $r$ from the Foreground Reliability Gate (Sec.~\ref{sec3.2}), \texttt{FVG-PT} fine-tunes the Foreground Distillation Compensation module (Sec.~\ref{sec3.3}) to focus visual attention on the foreground. Meanwhile, the independent Prior Calibration module (Sec.~\ref{sec3.4}) is simultaneously fine-tuned to address the BNT problem.

\subsection{Preliminaries}  \label{sec3.1}
Identical to other prompt tuning backbone models, \texttt{FVG-PT} builds on the frozen CLIP as the foundation VLM, which consists of a pretrained ViT-based \cite{dosovitskiy2020vit} visual encoder $f(\cdot)$ and a text encoder $g(\cdot)$. To enable efficient adaptation to downstream tasks, prompt tuning introduces a set of learnable prompt vectors of length $L$, which are appended to the visual input $x$, text input $t$, or intermediate layers of the encoder:
\begin{align}
    \boldsymbol{P}=[\mathrm{\theta}]_{1}[\mathrm{\theta}]_{2}\ldots[\mathrm{\theta}]_{L}
\end{align}
The text prompt $\boldsymbol{P}_t$ is typically constructed by concatenating a learnable prompt $\boldsymbol{P}$ with the [CLASS] token covering all candidate classes $\mathbf{C}=\left\{t_{i}\right\}_{i=1}^{n}$. For visual prompts, the prompt vectors are normally organized as the prefix of the patch tokens of the image $x$, forming the sequence $(\boldsymbol{P}_v,x)$. Prompts applied within intermediate encoder layers follow a similar pattern. To optimize the predicted output of the classification head, in image-text alignment stage, prompt tuning updates the parameters of the learnable prompts using a cross-entropy loss:
\begin{align}
    \mathcal{L}_{\mathrm{CE}}=-\sum_{i} {h}_{i} \log \frac{\exp \left(sim(g({\boldsymbol{P}_t}_y), f(\boldsymbol{P}_v,x)) / \tau\right)}{\sum_{i=1}^{n} \exp \left(sim(g(\boldsymbol{P}_{t_i}), f(\boldsymbol{P}_v,x))/ \tau\right)}
\end{align}
where $h_i$ is the one-hot label corresponding to the  $i$-th candidate in the set $\mathbf{C}$, and $\text{sim}(\cdot, \cdot)$ denotes the cosine similarity. The term $\tau$ is a predefined temperature parameter in CLIP. Further details on the workflow of the backbone models used in \texttt{FVG-PT} are provided in \textit{Supplementary Material~\ref{sec:A4}}.

\subsection{Foreground Reliability Gate}  \label{sec3.2}
Motivated by the empirical findings in Sec.~\ref{sec:intro} that suggest guiding visual attention toward the foreground, \texttt{FVG-PT} constructs a high-quality explicit foreground to serve as supervision for attention guidance.

Specifically, inspired by DAPT \cite{zhang2025dapt}, we employ a pretrained SEEM \cite{zou2023seem} segmentation model to generate a foreground mask $m\in\{0,1\}^{H\times W}$ for the raw image $x$ (details and visualizations of SEEM are provided in \textit{Supplementary Material~\ref{sec:A2}}). This yields an initial foreground view $x^{\mathrm{fg}} = x \odot m$, where $\odot$ denotes the Hadamard product. To enable adaptive assessment, the Foreground Reliability Gate (FRG) introduces a two-layer MLP during base-class tuning that learns a dynamic weighting scheme between $x^{\mathrm{fg}}$ and $x$.

\paragraph{\bf Supervision.} FRG first leverages the backbone model to obtain hard supervision for reliability. Let $\mathbf{z}(x)\in\mathbb{R}^{C}$ denote the prediction logits over the base-class candidates $\mathbf{C}$ produced by the fully frozen backbone for input image $x$. FRG computes cross-entropy (CE) losses on both the full image and foreground view:
\begin{align}
    \ell_{\text {full }}=\operatorname{CE}(\mathbf{z}(x), t), \quad \ell_{\mathrm{fg}}=\operatorname{CE}\left(\mathbf{z}(x^{\mathrm{fg}}), t\right)
\end{align}
Clearly, if $\ell_{\mathrm{fg}}<\ell_{\mathrm{full}}$, the foreground view brings a lower classification loss, indicating higher reliability. Therefore, we construct a binary supervision target $r^* = \mathbb{I}\left[\ell_{\mathrm{fg}} < \ell_{\mathrm{full}}\right]\in\{0,1\}$ to supervise FRG during fine-tuning.

\paragraph{\bf Gating Mechanism.} To ensure plug-and-play compatibility, FRG constructs its input features based on the final logits (after image-text alignment) of the backbone. To improve the robustness of foreground quality assessment, we introduce $N_{\text{frg}}=3$ reliability indicators. Specifically, given the soft distributions $\mathbf{p}$ (computed with temperature $\tau_{d}$) over base classes for $x$ and $x^{\mathrm{fg}}$:
\begin{align}
    \mathbf{p}_{\text {full}}=\operatorname{softmax}\left({\mathbf{z}(x)}/{\tau_{d}}\right), \quad \mathbf{p}_{\mathrm{fg}}=\operatorname{softmax}\left({\mathbf{z}(x^{\mathrm{fg}})}/{\tau_{d}}\right)
\end{align}
FRG constructs a statistics vector $\mathbf{u}$ with dimension $N_{\text{frg}}=3$ as the input to the MLP, as illustrated in Fig.~\ref{Fig-2}(b):
\begin{align}
    \mathbf{u} = [\Delta H, \cos(\mathbf{p}_{\text{full}}, \mathbf{p}_{\text{fg}}), A(m)]
\end{align}
The first indicator is a \textit{distribution-based} criterion $\begin{aligned}\Delta H = H(\mathbf{p}_{\text{full}}) - H(\mathbf{p}_{\text{fg}})\end{aligned}$, where $H(\cdot)$ denotes the entropy of a probability distribution. If the foreground view is more reliable, $\mathbf{p}_{\text{fg}}$ becomes sharper and has lower entropy, leading to $\Delta H > 0$, indicating higher foreground confidence. The second indicator is a \textit{similarity-based} criterion, which constrains $\mathbf{p}_{\text{fg}}$ to remain close to $\mathbf{p}_{\text{full}}$, thereby reducing the risk of distributional shift caused by inaccurate masks. The third indicator is a \textit{geometric} criterion that measures the foreground-to-full area ratio, ensuring the foreground is sufficiently informative and not overly small.

During fine-tuning, FRG predicts a trust score $r=\text{sigmoid}(q) \in(0,1)$ based on output scalar logit $q=\text{MLP}(\mathbf{u})$. Since the supervision signal $r^{\star}$ is binary-labeled, we apply binary cross-entropy (BCE) loss to train FRG to identify which weighting mode of indicators in the input corresponds to reliable foreground view:
\begin{align}
    \mathcal{L}_{\mathrm{FRG}} = -r^{\star} \log r - (1 - r^{\star}) \log (1-r)
\end{align}

\subsection{Foreground Distillation Compensation}  \label{sec3.3}
FDC aims to guide visual attention toward the filtered foreground view, thereby correcting shifted attention. Since the alignment pattern of foreground guidance is different from original backbone, to avoid conflicts in optimization direction, FDC inserts a standard bottleneck adapter \cite{houlsby2019Adapter} after original feature alignment stage of backbones, as in Fig.~\ref{Fig-2}(a). Unlike common task-driven adapters (e.g., CLIP-Adapter \cite{gao2024clipadapter}), it is a compensation module that encourages foreground-oriented attention. It aligns the \textit{backbone image-text logits using foreground view} with \textit{compensated logits of full image $x$}, prompting the model to learn a feature re-projection that guides attention toward the foreground. Adapters are inserted into both visual and textual branches with separate parameters, thus avoiding disruption of cross-modal alignment due to unilateral fine-tuning.

Specifically, for the image and text features $f(x)$ and $g(t)$ output by the backbone, FDC introduces two-layer bottleneck MLPs (i.e., dimension of the hidden layers of MLPs are smaller than the input) with residual connections as adapters, producing re-projected features $\tilde{f}(x)$ and $\tilde{g}(t)$:
\begin{align}
    \tilde{f}(x)=\operatorname{L2Norm}(f(x)+\text{MLP}\left(f(x)\right)), \ \ \tilde{g}(t)=\operatorname{L2Norm}(g(t)+\text{MLP}\left(g(t)\right))
\end{align}
Then, through image-text interaction of the compensated features, we obtain the FDC logit $\mathbf{z}_{\text{FDC}}(x)$ and the predicted distribution $\mathbf{p}_{\text{FDC}}$:
\begin{align}
    \mathbf{z}_{\text{FDC}}(x) = [\tilde{f}(x)]^\top \tilde{g}(t), \quad \mathbf{p}_{\text {FDC}}=\operatorname{softmax}\left({\mathbf{z}_{\text{FDC}}(x)}/{\tau_{d}}\right)
\end{align}
During fine-tuning, FDC employs a KL divergence-based distillation objective. The target distribution is selected adaptively based on the trust score $r$. When the foreground is reliable ($r\to1$), FDC aligns more closely with the foreground distribution $\mathbf{p}_{\mathrm{fg}}$. Otherwise, it falls back to the full image distribution $\mathbf{p}_{\text {full}}$:
\begin{align}
    \label{eqn9}
    \mathcal{L}_{\text {dist}}=r \cdot \operatorname{KL}\left(\mathbf{p}_{\mathrm{fg}} \| \mathbf{p}_{\mathrm{FDC}}\right)+(1-r) \cdot \operatorname{KL}\left(\mathbf{p}_{\mathrm{full}} \| \mathbf{p}_{\mathrm{FDC}}\right)
\end{align}
The overall loss for the base branch combines the FRG loss $\mathcal{L}_{\mathrm{FRG}}$, the FDC cross-entropy loss, and the distillation loss weighted by $\lambda_{d}$.
\begin{align}
    \label{eqn10}
    \mathcal{L}_{\text {base}}=\operatorname{CE}(\mathbf{z}_{\text{FDC}}(x), t) + \mathcal{L}_{\text{FRG}} + \lambda_{d} \mathcal{L}_{\text {dist}}
\end{align}

\begin{figure}[t]
  \centering
  \includegraphics[width=\textwidth]{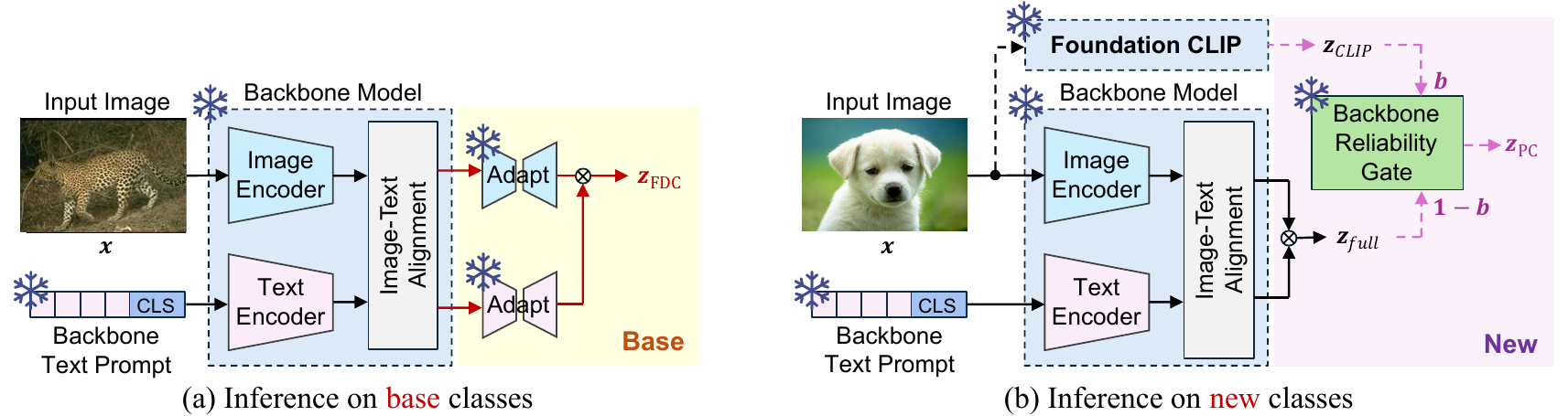}
  \caption{Inference stage of \texttt{FVG-PT} on (a) base branch and (b) new branch. The design of Prior Calibration (Sec.~\ref{sec3.4}) enables full decoupling between the two branches during inference, addressing the BNT problem.}
  \label{Fig-3}
\end{figure}

\subsection{Prior Calibration}  \label{sec3.4}
Since the foreground emphasis of FDC may reduce new-class generalization due to the BNT problem (empirical results are in Sec.~\ref{sec4.3}), the Prior Calibration (PC) module adopts a decoupled design at the logit level that fully separates the new branch from base (foreground) branch to avoid interference on the optimization direction. For optimizing the new branch, PC employs a Backbone Reliability Gate (BRG) to learn an adaptive balance with the CLIP prior.

\paragraph{\bf Decoupled Branches.} Although FDC and PC are both fine-tuned on the same base-class data, the logit-level design of \texttt{FVG-PT} allows complete decoupling between the base and new branches. As illustrated in Fig.~\ref{Fig-3}, during inference, the base branch uses the fine-tuned FDC adapters to compute the re-projected image-text logits $\mathbf{z}_{\text{FDC}}(x)$ for an input image $x$. For the new branch, PC first applies the original zero-shot CLIP and obtains prior logits for the same image using the initial text features $t_\text{CLIP}$ (e.g., ``\textit{A photo of a [CLASS]}''):
\begin{align}
\mathbf{z}_{\text{CLIP}}(x) = [{f^\text{CLIP}}(x)]^\top {g^\text{CLIP}}(t_\text{CLIP})
\end{align}
Next, PC learns a sample-adaptive CLIP trust score $b \in [0, 1]$ that controls the strength of CLIP prior on backbone and produces calibrated logits $\mathbf{z}_{\text{PC}}(x)$:
\begin{align}
\hat{y} = \arg \max_{c \in \mathbf{C}_\text{new}} \mathbf{z}_{\text{PC}}(x)_{c}, \quad \mathbf{z}_\text{PC}(x)= (1-b) \cdot \mathbf{z}_{\text {full}}(x)+ b \cdot \mathbf{z}_{\text{CLIP}}(x)
\end{align}
Under this design, the new branch does not call FDC, thus isolating it from base.

\paragraph{\bf Backbone Reliability Gate.} Similar to FRG (Sec.~\ref{sec3.2}), BRG introduces a learnable two-layer MLP that combines $N_{\text{brg}}=3$ logit-based statistics as an input vector $\mathbf{s}$, and outputs a predicted scalar $b= \text{sigmoid}(\text{MLP}(\mathbf{s}))$:
\begin{align}
    \mathbf{s} = \left[ H(\mathbf{p}_{\text{full}}),\, H(\mathbf{p}_{\text{CLIP}}),\,  \text{cos}(\mathbf{p}_{\text{full}}, \mathbf{p}_{\text{CLIP}}) \right] \\
\mathbf{p}_{\text {CLIP}}=\operatorname{softmax}\left({\mathbf{z}_{\text{CLIP}}(x)}/{\tau_{d}}\right) \quad \quad
\end{align}
During fine-tuning, BRG learns how to allocate weights across these statistics based on image-text feature distribution to achieve adaptive calibration. For example, low $H(\mathbf{p}_{\text{CLIP}})$ or high $H(\mathbf{p}_{\text{full}})$ indicate more confident CLIP prior (the distribution is sharper), therefore, the model is encouraged to assign a larger $b$, encouraging $\mathbf{z}_{\text{PC}}(x)$ to follow CLIP. The indicator $\text{cos}(\cdot, \cdot)$ constrains the similarity between the two distributions and drives $b$ toward the more confident one when the similarity is low.

To realize this behavior, BRG is trained with a joint objective that combines the cross-entropy loss for base-class discrimination and a KL regularization term that keeps the calibrated distribution close to the CLIP prior:
\begin{align}
\mathcal{L}_{\text {PC}}=\operatorname{CE}(\mathbf{z}_{\text{PC}}(x), t) + \operatorname{KL}\left(\mathbf{p}_{\text{CLIP}} \| \operatorname{softmax}\left({\mathbf{z}_{\text{PC}}(x)}/{\tau_{d}}\right)\right)
\end{align}
This objective learns the optimal trust score $b$ for each input sample, and learns the optimal mapping from $\mathbf{s}$ to $b$ on the entire training set.

\section{Experiments}  \label{sec4}

\begin{table}[t]
    \caption{Base-to-new generalization performance of 4 backbone models w/ or w/o our \texttt{FVG-PT} on 11 datasets. \texttt{FVG-PT} achieves performance improvements on both base-class fine-tuning and new-class generalization across all backbones.}
    \label{tab1}
    \centering
    \setlength\tabcolsep{2.7pt}

    \scalebox{0.8}{
\begin{tabular}{c|ccc|ccc|ccc|ccc} 
\toprule
\rowcolor{gray!10} {\cellcolor{gray!10}}                                  & \multicolumn{3}{c|}{\textbf{Average over 11}}                                                                                                                & \multicolumn{3}{c|}{\textbf{ImageNet}}           & \multicolumn{3}{c|}{\textbf{Caltech101}}         & \multicolumn{3}{c}{\textbf{Food101}}              \\
\rowcolor{gray!10} \multirow{-1.8}{*}{{\cellcolor{gray!10}}\textbf{Method}} & Base                                               & New                                                & HM                                                 & Base           & New            & HM             & Base           & New            & HM             & Base           & New            & HM              \\
\midrule
CoOp \cite{zhou2022coop}                                                 & 80.94                                              & 70.03                                              & 75.09                                              & 75.96          & 68.93          & 72.27          & 97.81          & 94.65          & 96.20          & \textbf{90.14} & 91.48          & \textbf{90.81}  \\
{\cellcolor{cyan!15}}\textbf{+FVG-PT}      & {\cellcolor{cyan!15}}\textbf{82.40}      & {\cellcolor{cyan!15}}\textbf{73.62}      & {\cellcolor{cyan!15}}\textbf{77.76}      & \textbf{76.65} & \textbf{70.20} & \textbf{73.28} & \textbf{98.13} & \textbf{94.76} & \textbf{96.42} & 89.43          & \textbf{91.74} & 90.57           \\ 
\midrule
KgCoOp \cite{yao2023kgcoop}                                              & 80.28                                              & 73.75                                              & 76.88                                              & 76.18          & 70.26          & 73.10          & 98.19          & 94.21          & 96.16          & \textbf{90.59} & \textbf{91.65} & \textbf{91.12}  \\
{\cellcolor{cyan!15}}\textbf{+FVG-PT} & {\cellcolor{cyan!15}}\textbf{82.06} & {\cellcolor{cyan!15}}\textbf{74.51} & {\cellcolor{cyan!15}}\textbf{78.10} & \textbf{77.14} & \textbf{70.28} & \textbf{73.55} & \textbf{98.26} & \textbf{94.76} & \textbf{96.48} & 90.07          & 91.63          & 90.84           \\ 
\midrule
PromptSRC \cite{khattak2023promptsrc}                                           & 80.81                                              & 73.98                                              & 77.24                                              & 77.00          & 70.60          & 73.66          & 98.00          & 94.32          & 96.12          & \textbf{90.80} & 91.74          & \textbf{91.27}  \\
{\cellcolor{cyan!15}}\textbf{+FVG-PT} & {\cellcolor{cyan!15}}\textbf{81.21} & {\cellcolor{cyan!15}}\textbf{74.82} & {\cellcolor{cyan!15}}\textbf{77.89} & \textbf{77.31} & \textbf{70.64} & \textbf{73.82} & \textbf{98.06} & \textbf{94.65} & \textbf{96.32} & 90.72          & \textbf{91.79} & 91.25           \\ 
\midrule
MMRL \cite{guo2025mmrl}                                                & 85.51                                              & 75.33                                              & 80.10                                              & 78.52          & \textbf{70.84} & 74.48          & 98.52          & \textbf{94.43} & 96.43          & \textbf{90.05} & 91.47          & \textbf{90.75}  \\
{\cellcolor{cyan!15}}\textbf{+FVG-PT}      & {\cellcolor{cyan!15}}\textbf{86.04}      & {\cellcolor{cyan!15}}\textbf{76.06}      & {\cellcolor{cyan!15}}\textbf{80.75}      & \textbf{78.55} & 70.82          & \textbf{74.48} & \textbf{98.77} & 94.32          & \textbf{96.49} & 89.76          & \textbf{91.54} & 90.64           \\ 
\hline\hline
\rowcolor{gray!10}                    & \multicolumn{3}{c|}{\textbf{StanfordCars}}                                                                                                                   & \multicolumn{3}{c|}{\textbf{OxfordPets}}         & \multicolumn{3}{c|}{\textbf{Flowers102}}         & \multicolumn{3}{c}{\textbf{DTD}}                  \\
\rowcolor{gray!10} \multirow{-1.8}{*}{{\cellcolor{gray!10}}\textbf{Method}}                   & Base                                               & New                                                & HM                                                 & Base           & New            & HM             & Base           & New            & HM             & Base           & New            & HM              \\ 
\midrule
CoOp \cite{zhou2022coop}                                                & 69.17                                              & 69.70                                              & 69.43                                              & \textbf{95.11} & 97.04          & \textbf{96.07} & 96.77          & 71.42          & 82.18          & 78.24          & 48.67          & 60.01           \\
\textbf{+FVG-PT}                                     & \textbf{73.99}                                     & \textbf{75.34}                                     & \textbf{74.66}                                     & 93.94          & \textbf{97.76} & 95.81          & \textbf{97.53} & \textbf{75.39} & \textbf{85.04} & \textbf{81.13} & \textbf{53.99} & \textbf{64.83}  \\ 
\midrule
KgCoOp \cite{yao2023kgcoop}                                              & 69.42                                              & 75.74                                              & 72.44                                              & \textbf{95.00} & 97.54          & 96.25          & 94.97          & 73.40          & 82.80          & 77.31          & \textbf{60.14} & 67.65           \\
\textbf{+FVG-PT}                                     & \textbf{74.56}                                     & \textbf{76.01}                                     & \textbf{75.28}                                     & 94.95          & \textbf{97.65} & \textbf{96.28} & \textbf{97.06} & \textbf{74.54} & \textbf{84.32} & \textbf{80.44} & 60.02          & \textbf{68.75}  \\ 
\midrule
PromptSRC \cite{khattak2023promptsrc}                                           & 72.26                                              & 75.27                                              & 73.73                                              & \textbf{95.53} & \textbf{97.43} & \textbf{96.47} & 95.54          & 73.48          & 83.07          & 79.63          & 61.23          & 69.23           \\
\textbf{+FVG-PT}                                     & \textbf{74.11}                                     & \textbf{75.98}                                     & \textbf{75.03}                                     & 95.37          & 97.20          & 96.28          & \textbf{96.58} & \textbf{74.47} & \textbf{84.10} & \textbf{81.71} & \textbf{63.41} & \textbf{71.41}  \\ 
\midrule
MMRL \cite{guo2025mmrl}                                                & 82.11                                              & 73.98                                              & 77.83                                              & 95.32          & 97.32          & 96.31          & 98.67          & 76.95          & 86.47          & 85.19          & 65.46          & 74.03           \\
\textbf{+FVG-PT}                                     & \textbf{83.53}                                     & \textbf{74.94}                                     & \textbf{79.00}                                     & \textbf{95.43} & \textbf{97.48} & \textbf{96.44} & \textbf{98.86} & \textbf{77.52} & \textbf{86.90} & \textbf{85.42} & \textbf{67.51} & \textbf{75.42}  \\ 
\hline\hline
\rowcolor{gray!10}                    & \multicolumn{3}{c|}{\textbf{EuroSAT}}                                                                                                                        & \multicolumn{3}{c|}{\textbf{FGVCAircraft}}       & \multicolumn{3}{c|}{\textbf{SUN397}}             & \multicolumn{3}{c}{\textbf{UCF101}}               \\
\rowcolor{gray!10} \multirow{-1.8}{*}{{\cellcolor{gray!10}}\textbf{Method}}                   & Base                                               & New                                                & HM                                                 & Base           & New            & HM             & Base           & New            & HM             & Base           & New            & HM              \\ 
\midrule
CoOp \cite{zhou2022coop}                                                & 88.43                                              & 45.87                                              & 60.41                                              & 35.59          & 31.01          & 33.14          & 80.21          & 73.84          & 76.89          & 82.94          & 77.72          & 80.25           \\
\textbf{+FVG-PT}                                     & \textbf{89.76}                                     & \textbf{58.28}                                     & \textbf{70.67}                                     & \textbf{39.38} & \textbf{35.45} & \textbf{37.31} & \textbf{81.55} & \textbf{77.26} & \textbf{79.35} & \textbf{84.90} & \textbf{79.66} & \textbf{82.20}  \\ 
\midrule
KgCoOp \cite{yao2023kgcoop}                                              & 82.60                                              & 60.38                                              & 69.76                                              & 34.51          & 35.39          & 34.94          & 80.46          & 76.57          & 78.47          & 83.82          & 75.99          & 79.71           \\
\textbf{+FVG-PT}                                     & \textbf{85.74}                                     & \textbf{65.00}                                     & \textbf{73.94}                                     & \textbf{36.79} & \textbf{35.87} & \textbf{36.32} & \textbf{82.04} & \textbf{76.84} & \textbf{79.35} & \textbf{85.63} & \textbf{76.96} & \textbf{81.06}  \\ 
\midrule
PromptSRC \cite{khattak2023promptsrc}                                           & \textbf{79.12}                                     & 58.38                                              & 67.19                                              & 35.95          & 35.39          & 35.67          & 81.93          & 77.86          & 79.84          & 83.14          & 78.10          & 80.54           \\
\textbf{+FVG-PT}                                     & 76.43                                              & \textbf{61.36}                                     & \textbf{68.07}                                     & \textbf{36.61} & \textbf{36.47} & \textbf{36.54} & \textbf{82.50} & \textbf{78.13} & \textbf{80.26} & \textbf{83.92} & \textbf{78.96} & \textbf{81.36}  \\ 
\midrule
MMRL \cite{guo2025mmrl}                                                & 95.98                                              & 64.08                                              & 76.85                                              & 45.44          & 35.03          & 39.56          & \textbf{82.69} & 79.21          & 80.91          & 88.11          & 79.88          & 83.79           \\
\textbf{+FVG-PT}                                     & \textbf{96.31}                                     & \textbf{65.38}                                     & \textbf{77.89}                                     & \textbf{48.56} & \textbf{36.71} & \textbf{41.81} & 82.53          & \textbf{79.57} & \textbf{81.02} & \textbf{88.73} & \textbf{80.91} & \textbf{84.64}  \\
\bottomrule
\end{tabular}
}
\end{table}

\subsection{Experimental Setup}  \label{sec4.1}

\paragraph{\bf Datasets.} Following mainstream prompt tuning practice \cite{zhou2022coop, khattak2023promptsrc}, \texttt{FVG-PT} is evaluated by base-to-new generalization and cross-dataset transfer tasks on 11 datasets with diverse data distributions, containing ImageNet \cite{deng2009imagenet}, Caltech101 \cite{fei2004caltech}, Food101 \cite{bossard2014food}, StanfordCars \cite{krause2013cars}, OxfordPets \cite{parkhi2012pets}, Flowers102 \cite{nilsback2008flowers}, DTD \cite{cimpoi2014dtd}, EuroSAT \cite{helber2019eurosat}, FGVCAircraft \cite{maji2013aircraft}, SUN397 \cite{xiao2010sun}, and UCF101 \cite{soomro2012ucf101}. Crucially, to avoid \textit{data leakage}, \texttt{FVG-PT} uses exactly the same training data as the backbone models during fine-tuning.

\paragraph{\bf Backbone Models.}  To verify the plug-and-play property of \texttt{FVG-PT}, we select 4 backbone models with different forms of learnable prompts as baselines. These include CoOp \cite{zhou2022coop} with text prompts, PromptSRC \cite{khattak2023promptsrc} with cross-modal prompts, KgCoOp \cite{yao2023kgcoop} with external knowledge, and MMRL \cite{guo2025mmrl} with mid-layer encoder plugins. Pipelines of the backbone models are in \textit{Supplementary Material~\ref{sec:A4}}. We also compare with prompt tuning methods that consider the attention pattern of the visual encoder, namely ProGrad \cite{zhu2023prograd} and DAPT-S \cite{zhang2025dapt}. Furthermore, we report additional baseline comparisons in \textit{Supplementary Material~\ref{sec:B2}}.

\begin{table}[t]
    \caption{Cross-Dataset Transfer performance of 4 backbone models w/ or w/o our \texttt{FVG-PT} on 11 datasets. \texttt{FVG-PT} achieves performance improvements on both the ImageNet source and multiple target datasets across all backbones.}
    \label{tab2}
    \centering
    \setlength\tabcolsep{1.8pt}

    \scalebox{0.8}{

\begin{tabular}{c|c|ccccccccccc} 
\toprule
\rowcolor{gray!10} {\cellcolor{gray!10}}                                  & \textbf{Source}                                    & \multicolumn{11}{c}{\textbf{Target}}                                                                                                                                                                                          \\
\rowcolor{gray!10} \multirow{-2}{*}{{\cellcolor{gray!10}}\textbf{Method}} & ImgNet                                           & \textbf{Avg.}                                               & Caltech        & Pets           & Cars           & Flower         & Food           & Aircraft       & SUN            & DTD            & EuroSAT        & UCF             \\ 
\midrule
CoOp                                                                                                    & 71.55                                              & 64.57                                              & 93.67          & \textbf{90.11} & 64.23          & 70.00          & 86.52          & 22.11          & 65.54          & 43.97          & 41.49          & 68.07           \\
{\cellcolor{cyan!15}}\textbf{+FVG-PT}                                                       & {\cellcolor{cyan!15}}\textbf{72.09}    & {\cellcolor{cyan!15}}\textbf{65.72}    & \textbf{94.16} & 89.75          & \textbf{65.56} & \textbf{71.66} & \textbf{86.77} & \textbf{24.75} & \textbf{66.54} & \textbf{45.39} & \textbf{43.77} & \textbf{68.89}  \\ 
\midrule
KgCoOp                                                                                                  & 71.20                                              & 65.41                                              & 93.96          & \textbf{90.49} & 65.22          & 71.05          & 86.37          & 23.85          & 66.86          & 46.39          & 41.41          & 68.49           \\
{\cellcolor{cyan!15}}\textbf{+FVG-PT}                                                    & {\cellcolor{cyan!15}}\textbf{72.17} & {\cellcolor{cyan!15}}\textbf{65.82} & \textbf{94.04} & 90.27          & \textbf{65.58} & \textbf{71.25} & \textbf{86.55} & \textbf{24.93} & \textbf{66.93} & \textbf{46.63} & \textbf{43.24} & \textbf{68.81}  \\ 
\midrule
PromptSRC                                                                                               & 72.43                                              & 64.76                                              & \textbf{92.13} & 90.90          & 65.34          & \textbf{68.57} & 86.12          & 23.46          & 66.80          & \textbf{45.04} & 41.25          & 67.94           \\
{\cellcolor{cyan!15}}\textbf{+FVG-PT}                                                    & {\cellcolor{cyan!15}}\textbf{72.86} & {\cellcolor{cyan!15}}\textbf{65.03} & 92.29          & \textbf{90.54} & \textbf{65.73} & 68.53          & \textbf{86.14} & \textbf{23.55} & \textbf{66.86} & 44.86          & \textbf{43.10} & \textbf{68.65}  \\ 
\midrule
MMRL                                                                                                    & 73.70                                              & 66.47                                              & \textbf{94.20} & \textbf{91.63} & 65.78          & 71.74          & 85.06          & 24.78          & 67.26          & 43.79          & 51.95          & 68.54           \\
{\cellcolor{cyan!15}}\textbf{+FVG-PT}                                                    & {\cellcolor{cyan!15}}\textbf{73.91} & {\cellcolor{cyan!15}}\textbf{66.99} & 94.14          & 91.14          & \textbf{66.46} & \textbf{72.15} & \textbf{86.15} & \textbf{25.44} & \textbf{67.42} & \textbf{44.74} & \textbf{53.43} & \textbf{68.81}  \\
\bottomrule
\end{tabular}
}
\end{table}

\paragraph{\bf Implementation Details.}  \texttt{FVG-PT} is fine-tuned with exactly the same input data, initialization, and hyperparameters as the backbone models. For a fair comparison, all methods adopt a 16-shot setting with $ep=10$ training epochs and learning rate $lr=0.0035$. For \texttt{FVG-PT}, we set the hidden dimension of the FDC bottleneck adapters to ${dim}_{\text{FDC}}=64$, the hidden dimension of both Reliability Gate MLPs to ${dim}_{\text{RG}}=32$, and the temperature coefficient to $\tau_{d}=2.0$. Based on the ablation results in Sec.~\ref{sec4.3}, we choose the weight of the FDC distillation loss $\mathcal{L}_{\text {dist}}$ to $\lambda_{d}=10.0$. Further details of the experimental settings are provided in \textit{Supplementary Material~\ref{sec:A}}.

\subsection{Experimental Results}  \label{sec4.2}

\paragraph{\bf Base-to-New Generalization.}  Following mainstream prompt tuning practice \cite{zhou2022coop, khattak2023maple, guo2025mmrl}, each dataset splits its categories evenly into base and new classes. To prevent data leakage, we ensure that (1) \texttt{FVG-PT} and the backbone model use exactly the same base-class data for fine-tuning, and (2) the model only accesses the base training set during the fine-tuning stage, while inference is performed on both base and new classes. HM denotes the harmonic mean of the accuracies on base and new tasks. As reported in Tab.~\ref{tab1}, \texttt{FVG-PT} achieves consistent improvements on both base adaptation and new-class generalization across various forms of backbone models. Particularly, the logit-level plug-and-play structure of \texttt{FVG-PT} overcomes the difficulty that previous plug-and-play methods \cite{zhang2024dept, li2025dpc, li2025augpt} face when adapting to prompt tuning with mid-layer encoder plugins \cite{seputis2024mma, guo2025mmrl}, resulting in the improvements based on the state-of-the-art MMRL \cite{guo2025mmrl} on multiple datasets. Error bar analysis is provided in \textit{Supplementary Material~\ref{sec:B1}}.

\paragraph{\bf Cross-Dataset Transfer.}  To evaluate the generalization ability of prompt tuning across data distributions, we perform cross-dataset transfer by fine-tuning on all categories of ImageNet as the source dataset, and then conducting zero-shot inference on the remaining target datasets described in Sec.~\ref{sec4.1} using the fine-tuned model. As shown in Tab.~\ref{tab2}, \texttt{FVG-PT} improves the performance of all 4 backbones on the ImageNet source, indicating the effectiveness of foreground attention guidance. Moreover, \texttt{FVG-PT} achieves clear zero-shot gains on most target datasets, which suggests that the weight allocation scheme learned by Prior Calibration (Sec.~\ref{sec3.4}) generalizes well across different data distributions.

\begin{figure}[t]
  \centering
  \begin{subfigure}{0.61\linewidth}
    \includegraphics[width=\textwidth]{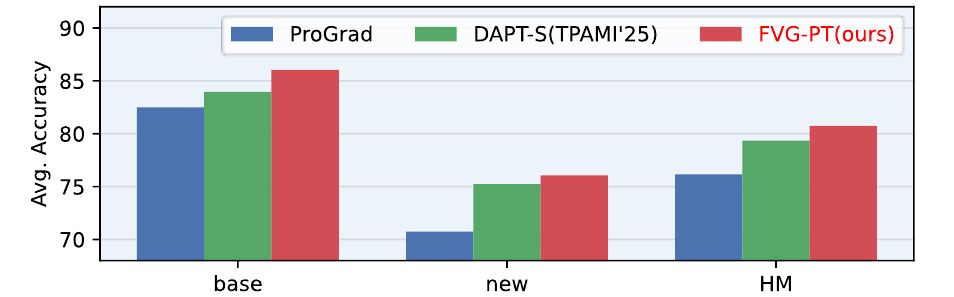}
    \caption{Visual attention-related methods}
    \label{Fig-4a}
  \end{subfigure}
  \hfill
  \begin{subfigure}{0.33\linewidth}
    \includegraphics[width=\textwidth]{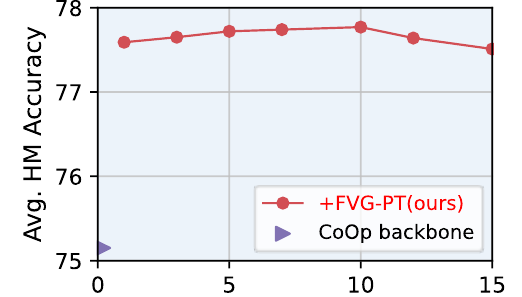}
    \caption{$\lambda_{d}$}
    \label{Fig-4b}
  \end{subfigure}
  \caption{Base-to-new performance comparison across (a) visual attention-related prompt tuning methods and (b) different weights of the FDC distillation loss $\lambda_{d}$.}
  \label{Fig-4}
\end{figure}

\paragraph{\bf Compare with Other Related Methods.} To further verify the effectiveness of the foreground attention guidance in \texttt{FVG-PT}, we compare its HM performance with two prompt tuning methods that also consider visual attention in Fig.~\ref{Fig-4}(a), namely ProGrad \cite{zhu2023prograd}, which implicitly regulates attention by constraining gradient propagation, and DAPT-S \cite{zhang2025dapt}, which directly uses all foreground cues for optimization. It can be observed that \texttt{FVG-PT} achieves better overall performance than the others, indicating that (1) explicit foreground supervision provides more effective control over visual attention, and (2) the combination of Foreground Reliability Gate (Sec.~\ref{sec3.2}) and Foreground Distillation Compensation (Sec.~\ref{sec3.3}) guides attention toward the correct foreground regions more reliably. In addition, benefiting from its plug-and-play property, \texttt{FVG-PT} has the potential to further improve performance when combined with stronger backbone models.

\begin{table}[t]
    \caption{Ablation study of components in \texttt{FVG-PT} on base-to-new tasks. Detailed ablation of losses and segmentation module are in \textit{Supplementary Material~\ref{sec:B3}}.}
    \label{tab3}
    \centering
    \setlength\tabcolsep{6pt}
    \scalebox{0.8}{
\begin{tabular}{cccccc|ccc|c} 
\toprule
\rowcolor{gray!10} {\cellcolor{gray!10}}                   & \textbf{FRG}                                                      & \multicolumn{2}{c}{\textbf{FDC}}                                                                                                      & \multicolumn{2}{c|}{\textbf{PC}}                                                                                                   & \multicolumn{3}{c|}{\textbf{Average over 11}} & {\cellcolor{gray!10}}                                                      \\
\rowcolor{gray!10} \multirow{-2}{*}{{\cellcolor{gray!10}}} & $\mathcal{L}_{\text {FRG}}$ & $\mathcal{L}_{\text{CE}}$ & $\mathcal{L}_{\text {dist}}$ & $\mathcal{L}_{\text {CE}}$ & $\mathcal{L}_{\text {KL}}$ & Base  & New   & HM                            & \multirow{-2}{*}{{\cellcolor{gray!10}}\textbf{$\Delta$}}  \\ 
\midrule
                                                                                         & \ding{55}                                        & \ding{55}                                       & \ding{55}                                         & \ding{55}                                       & \ding{55}                                       & 80.94 & 70.03 & 75.09                         & (baseline)                                                                                          \\
(1)                                                                                      & \ding{55}                                        & \ding{51}                                       & \ding{51}                                         & \ding{55}                                       & \ding{55}                                       & 81.04 & 63.14 & 70.98                         & \textcolor{X}{$-$ 4.11}                                                                                     \\
(2)                                                                                      & \ding{55}                                        & \ding{55}                                       & \ding{55}                                         & \ding{51}                                       & \ding{51}                                       & 80.94 & 73.62 & 77.11                         & \textcolor{V}{$+$ 2.02}                                                                                    \\
(3)                                                                                      & \ding{51}                                        & \ding{51}                                       & \ding{51}                                         & \ding{55}                                       & \ding{55}                                       & 82.40 & 63.45 & 71.69                         & \textcolor{X}{$-$ 3.40}                                                                                     \\
(4)                                                                                      & \ding{55}                                        & \ding{51}                                       & \ding{51}                                         & \ding{51}                                       & \ding{51}                                       & 81.04 & 73.62 & 77.15                         & \textcolor{V}{$+$ 2.06}                                                                                    \\
(5)                                                                                      & \ding{51}                                        & \ding{51}                                       & \ding{51}                                         & \ding{51}                                       & \ding{51}                                       & 82.40 & 73.62 & 77.76                         & \textcolor{V}{$+$ 2.67}                                                                                    \\
\bottomrule
\end{tabular}
}
\end{table}

\subsection{Ablation Study}  \label{sec4.3}
In this section, we discuss the impact of each \texttt{FVG-PT} sub-module, the weight of the distillation loss $\lambda_d$, and the construction of learnable layers and gating indicators on model performance. The evaluation is based on base-to-new tasks of CoOp. Further experimental details can be found in \textit{Supplementary Material~\ref{sec:B}}.

\paragraph{\bf Validity of Proposed Components.}  Tab.~\ref{tab3} presents the contributions of each sub-module. We observe that (1) introducing \textbf{FDC} (Foreground Distillation Compensation, Sec.~\ref{sec3.3}) and (3) introducing both FDC and \textbf{FRG} (Foreground Reliability Gate, Sec.~\ref{sec3.2}) consistently improve base performance, which indicates that guiding visual attention toward the foreground benefits the target task. Meanwhile, controlling foreground quality is necessary, and (4) full model without FRG further confirms this point. However, both configurations (1) and (3) lack \textbf{PC} (Prior Calibration, Sec.~\ref{sec3.4}), leading to a clear drop in new-class performance, revealing a strong BNT effect. In contrast, (4) and (5) that include PC achieve significant gains on new, which shows that PC effectively mitigates BNT. The configuration of (2) enabling only PC module also demonstrates that PC fully decouples the base and new branches, eliminating the interference of base class fine-tuning on generalization. More detailed ablations in \textit{Supplementary Material~\ref{sec:B3}} show that each loss term contributes positively to \texttt{FVG-PT}.

\paragraph{\bf Impact of FDC Distillation Loss.}  Since \texttt{FVG-PT} introduces an adaptive design into most of its components, we mainly examine the impact of the distillation loss weight $\lambda_d$ (Eqn.~\ref{eqn10}) in FDC. Fig.~\ref{Fig-4}(b) shows that \texttt{FVG-PT} achieves the best HM performance when $\lambda_d=10$. However, it is worth noting that even without adjusting $\lambda_d$ (for example, simply setting $\lambda_d=1$), \texttt{FVG-PT+CoOp} still significantly outperforms the CoOp backbone. This result suggests that \texttt{FVG-PT} can achieve \textbf{fully adaptive} behavior with only a small performance cost.

\begin{table}[t]
    \centering
    
	\begin{minipage}{0.5\linewidth}
		\centering
        \caption{HM Performance under different hidden layer dimensions.}
        \label{tab4}
        \setlength\tabcolsep{4pt}
        \scalebox{0.8}{
            \begin{tabular}{ccccc} 
            \toprule
            \rowcolor{gray!10} {\cellcolor{gray!10}}                                   & \multicolumn{4}{c}{\textbf{Hidden layer dimension}}  \\
            \rowcolor{gray!10} \multirow{-2}{*}{{\cellcolor{gray!10}}\textbf{Modules}} & 16    & 32    & 64    & 128                          \\ 
            \midrule
            Adapters in FDC                                                                                    & 77.50 & 77.53 & \textbf{77.76} & 77.62                        \\
            MLPs in RG                                                                                         & 77.68 & \textbf{77.76} & 77.58 & 77.63                        \\
            \bottomrule
            \end{tabular}
        }
	\end{minipage}
    \hfill
	\begin{minipage}{0.45\linewidth}
            \centering
            \caption{Performance under different numbers of layers.}
            \label{tab5}
            \setlength\tabcolsep{4pt}
            \scalebox{0.8}{
        \begin{tabular}{cccc} 
\toprule
\rowcolor{gray!10} {\cellcolor{gray!10}}                                                                                                                                 & \multicolumn{3}{c}{\textbf{Average over 11}}  \\
\rowcolor{gray!10} \multirow{-2}{*}{{\cellcolor{gray!10}}\begin{tabular}[c]{@{}>{\cellcolor{gray!10}}c@{}}\textbf{Layers of}\\\textbf{MLPs}\end{tabular}} & Base  & New   & HM \\ 
\midrule
1 & 81.90 & 73.51 & 77.48 \\
2 & \textbf{82.40} & \textbf{73.62} & \textbf{77.76} \\
3 & 82.28 & 73.57 & 77.68 \\
\bottomrule
\end{tabular}
        }
	\end{minipage}
    
\end{table}

\begin{table}[t]
    \caption{Ablation study of indicators in \texttt{FVG-PT} on base-to-new tasks. Detailed ablation of losses and segmentation module are in \textit{Supplementary Material~\ref{sec:B3}}.}
    \label{tab6}
    \centering
    \setlength\tabcolsep{4.5pt}
    \scalebox{0.8}{
\begin{tabular}{cccccc|ccc|c} 
\toprule
\rowcolor{gray!10} {\cellcolor{gray!10}}                   & \multicolumn{3}{c}{\textbf{FRG indicators}} & \multicolumn{2}{c|}{\textbf{BRG indicators}} & \multicolumn{3}{c|}{\textbf{Average over 11}} & {\cellcolor{gray!10}}                                  \\
\rowcolor{gray!10} \multirow{-2}{*}{{\cellcolor{gray!10}}} & Distribution & Similarity & Geometric        & Distribution & Similarity                    & Base  & New   & HM                            & \multirow{-2}{*}{{\cellcolor{gray!10}}\textbf{$\Delta$}}  \\ 
\midrule
                                                                                   & \ding{51}   & \ding{51} & \ding{51}      & \ding{51}   & \ding{51}                    & 82.40 & 73.62 & 77.76                         & (full)                                                             \\
(1)                                                                                & \ding{55}   & \ding{51} & \ding{51}      & \ding{51}   & \ding{51}                    & 82.18 & 73.62 & 77.66                         & \textcolor{X}{$-$ 0.10}                                                             \\
(2)                                                                                & \ding{51}   & \ding{55} & \ding{51}      & \ding{51}   & \ding{51}                    & 82.23 & 73.62 & 77.69                         & \textcolor{X}{$-$ 0.07}                                                             \\
(3)                                                                                & \ding{51}   & \ding{51} & \ding{55}      & \ding{51}   & \ding{51}                    & 82.32 & 73.62 & 77.73                         & \textcolor{X}{$-$ 0.03}                                                            \\
(4)                                                                                & \ding{51}   & \ding{51} & \ding{51}      & \ding{55}   & \ding{51}                    & 82.40 & 73.53 & 77.71                         & \textcolor{X}{$-$ 0.05}                                                             \\
(5)                                                                                & \ding{51}   & \ding{51} & \ding{51}      & \ding{51}   & \ding{55}                    & 82.40 & 73.33 & 77.60                         & \textcolor{X}{$-$ 0.16}                                                             \\
\bottomrule
\end{tabular}
}
\end{table}

\paragraph{\bf Influence of Hidden Layer Dimension.}  In this section, we study how different values of hidden dimension at initialization affect the performance of \texttt{FVG-PT}. As shown in Tab.~\ref{tab4}, \texttt{FVG-PT} achieves the highest HM score when the hidden dimension of the FDC adapters is set to ${dim}_{\text{FDC}}=64$ and the hidden dimension of the MLPs in Reliability Gates (RGs) is set to ${dim}_{\text{RG}}=32$. However, overall, the performance does not decrease significantly under other dimensional settings. We hypothesize that this is because prompt tuning is performed on small scale few-shot data, so relatively small hidden layers are already sufficient to model the modest foreground feature compensation in FDC and the mapping from statistics to scalar weights in RGs.

\paragraph{\bf Different MLP Depths in Reliability Gates.}  We further examine how the number of MLP layers in the RGs affects the performance of \texttt{FVG-PT}. Results in Tab.~\ref{tab5} show that a single-layer MLP (without a hidden layer) causes a slight drop in HM, while the performances of 2-layer and 3-layer MLPs are very similar. This observation suggests that the RGs mainly learn an approximately monotonic reliability mapping in the statistics space, so the decision function has low complexity and does not rely on strong nonlinear representations in hidden layers. In summary, the RGs are insensitive to the capacity of the MLP.

\begin{table}[t]
    \caption{Computation cost of MMRL \cite{guo2025mmrl} backbone and \texttt{FVG-PT} on Flowers102 dataset.}
    \label{tab7}
    \centering
    \setlength\tabcolsep{4pt}
    \scalebox{0.8}{
\begin{tabular}{cc|cccc} 
\toprule
\rowcolor{gray!10} \textbf{Stage} & \textbf{Learnable Params} & \textbf{Tuning FPS}$\uparrow$ & \textbf{GPU Memory}$\downarrow$ & \textbf{Inference FPS}$\uparrow$ & \textbf{HM}$\uparrow$  \\ 
\midrule
MMRL                                       & 4.99 M                    & 87.52               & 2009.9 MB             & 610.6                  & 80.10        \\
\textbf{FVG-PT}                           & 0.13 M                   & 153.36               & 808.4 MB             & 847.6                  & 80.75        \\
\bottomrule
\end{tabular}
}
\end{table}

\paragraph{\bf Effect of Indicators in Reliability Gates.}  In Tab.~\ref{tab6}, we analyze the contribution of each indicator introduced in the RGs by removing exactly one indicator at a time from the full \texttt{FVG-PT} model, and measuring the performance drop while keeping other indicators unchanged. We observe that removing any single indicator reduces the HM score, indicating that all of them play a positive role in constraining the foreground view quality and the CLIP prior. Meanwhile, the performance degradation remains small, suggesting that the RGs are robust and do not heavily rely on any individual indicator.

\paragraph{\bf Computational Cost.}  For prompt tuning backbones with ViT-B/16 encoders, \texttt{FVG-PT} activates only 0.13M trainable parameters during fine-tuning, which is more parameter-efficient than advanced baseline models. As reported in Tab.~\ref{tab7}, compared with the backbone, \texttt{FVG-PT} achieves advantages in terms of parameter count, memory cost, and processing FPS (Frame Per Second) during fine-tuning.

\section{Conclusion}
We present \texttt{FVG-PT}, a plug-and-play prompt tuning enhancement method that exploits explicit foreground supervision. We trace the failure modes of existing prompt tuning to shifts in attention away from the foreground. To correct this shift, built on pre-tuned backbones, \texttt{FVG-PT} uses model-agnostic Foreground Distillation Compensation to guide visual attention under the supervision of foreground view, while a Foreground Reliability Gate is introduced to maintain the view quality. Meanwhile, to avoid BNT, Prior Calibration fully decouples the new branch from the base, and learns to balance the new branch with the CLIP prior via a Backbone Reliability Gate. During fine-tuning, the two Reliability Gates learn adaptive weighting patterns that allow the model to assess both the foreground view or the CLIP prior adaptively. Experiments confirm that \texttt{FVG-PT} is compatible with diverse backbones and remains consistent and robust across different hyperparameter settings, which validates its adaptive behavior.
\paragraph{\bf Limitation and Future Work.}  Although \texttt{FVG-PT} is compatible with most prompt tuning methods (especially mid-layer plugin approaches), a few exceptions remain. \texttt{FVG-PT} is not suitable for visual-only prompt tuning methods that lack a text branch (e.g., VPT \cite{jia2022vpt}). In addition, for knowledge distillation methods that use all unlabeled images in the dataset (e.g., PromptKD \cite{li2024promptkd}), the cost of obtaining foreground views increases significantly. As future work, we plan to further improve the adaptability of \texttt{FVG-PT}, for instance, by developing self-supervised variants that can better adapt to arbitrary backbones.


%
%
\bibliographystyle{splncs04}
\bibliography{main}

\input{sup/sup_mat.tex}

\end{document}

%% file: sup/sup_mat.tex
\clearpage
\setcounter{section}{0}
\setcounter{subsection}{0}

\appendix

\begin{center}
    {\Large \textbf{FVG-PT: Adaptive Foreground View-Guided Prompt Tuning for Vision-Language Models}}\\[1em]
    {\large \textit{Supplementary Material}}
\end{center}

\texttt{FVG-PT} is a foreground-guided plug-and-play prompt tuning enhancement method. To adaptively focus the attention of visual encoder on the image foreground, \texttt{FVG-PT} employs a \textit{Foreground Reliability Gate} (FRG) to improve the quality of foreground supervision, a \textit{Foreground Distillation Compensation} (FDC) module built on backbone models to control the focus of visual attention, and a \textit{Prior Calibration} (PC) module to mitigate the BNT problem.

In supplementary material, we provide further details on \texttt{FVG-PT}, including:
\begin{itemize}
    \item \textbf{Appendix~\ref{sec:B4}}: quantitative analysis and additional case studies of foreground attention shift in prompt tuning.
    \item \textbf{Appendix~\ref{sec:A}}: more implementation details, e.g., configurations of the SEEM semantic segmentation model and backbones, to support reproducibility.
    \item \textbf{Appendix~\ref{sec:B}}: additional experiments, including error bar analysis, data efficiency analysis, and more detailed ablations.
    \item \textbf{Appendix~\ref{sec:C}}: bad case studies that further analyze the limitations of \texttt{FVG-PT} and potential directions for improvement.
\end{itemize}

\section{Details of Shifts in Foreground Attention} \label{sec:B4}

In Sec.~\ref{sec:intro} of the main text, we use attention map-based case studies in Fig.~\ref{Fig-1} to attribute the failure modes of prompt tuning to an attention shift of the visual encoder away from the image foreground. In this section, we further validate this conclusion through quantitative analysis and additional case studies.

\subsection{Quantitative Analysis}

\begin{figure}[b]
  \centering
  \begin{subfigure}{0.45\linewidth}
    \includegraphics[width=\textwidth]{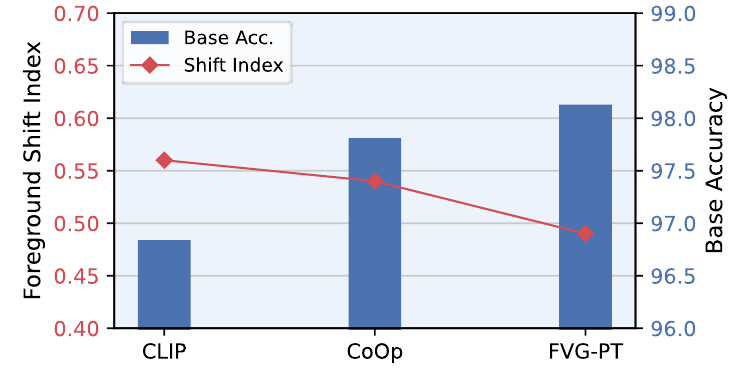}
    \caption{Caltech101 dataset}
    \label{Fig-s1a}
  \end{subfigure}
  \hfill
  \begin{subfigure}{0.45\linewidth}
    \includegraphics[width=\textwidth]{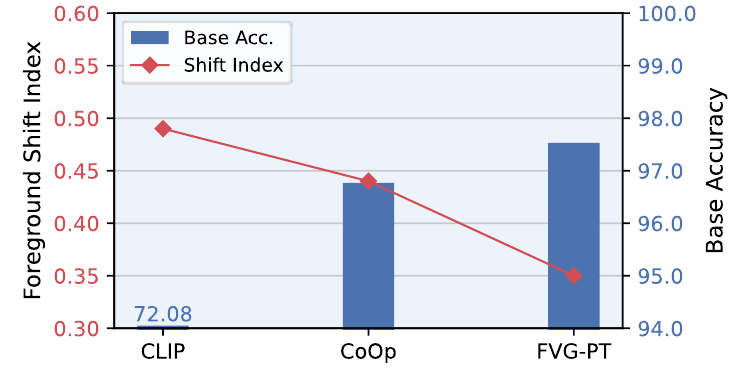}
    \caption{Flowers102 dataset}
    \label{Fig-s1b}
  \end{subfigure}
  \caption{Trends of the foreground shift index and base-class accuracy for CLIP, fine-tuned CoOp, and our \texttt{FVG-PT} on the (a) Caltech101 and (b) Flowers102 datasets.}
  \label{Fig-S1}
\end{figure}

\begin{figure}[t]
  \centering
  \includegraphics[width=0.95\textwidth]{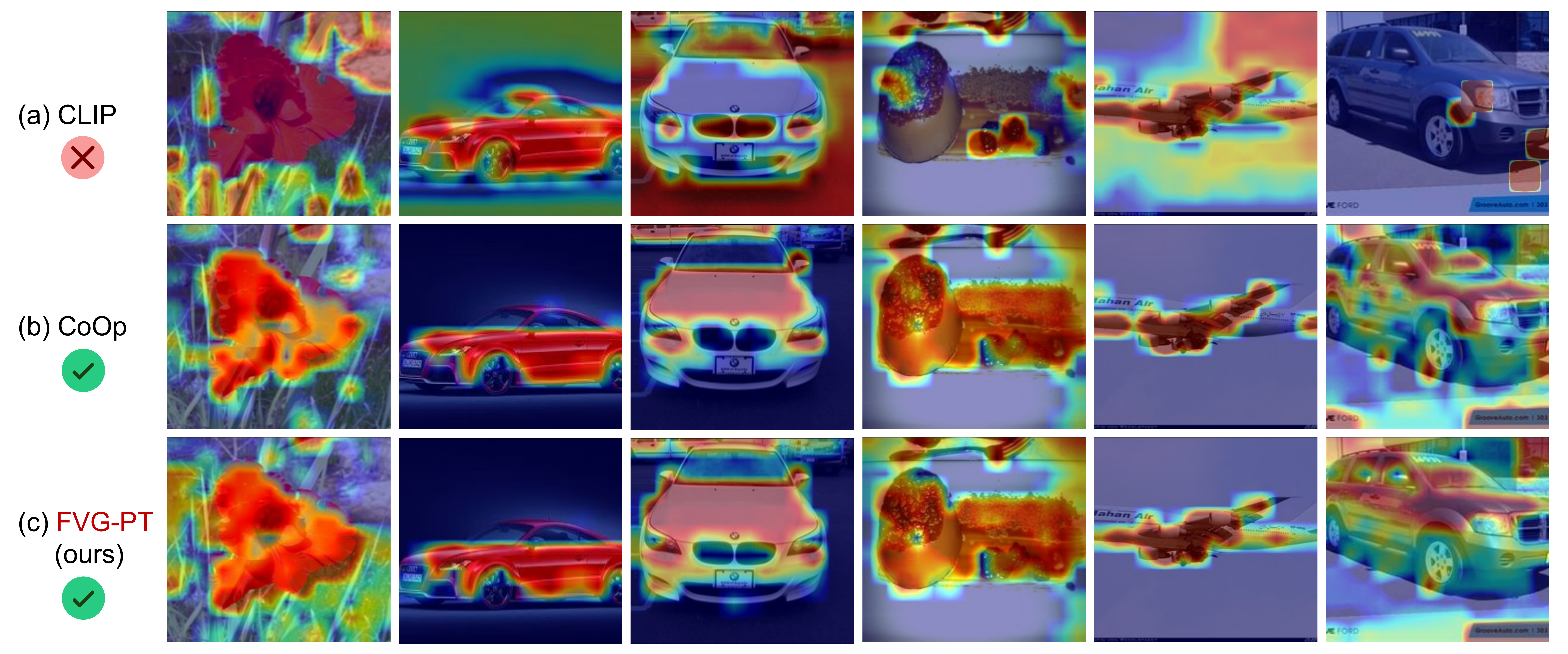}
  \caption{Failure cases of (a) the original CLIP compared with (b) CoOp and (c) our \texttt{FVG-PT}. Attention maps of visual encoders are generated by Grad-CAM \cite{selvaraju2017gradcam}.}
  \label{Fig-S2}
\end{figure}
\begin{figure}[t]
  \centering
  \includegraphics[width=0.95\textwidth]{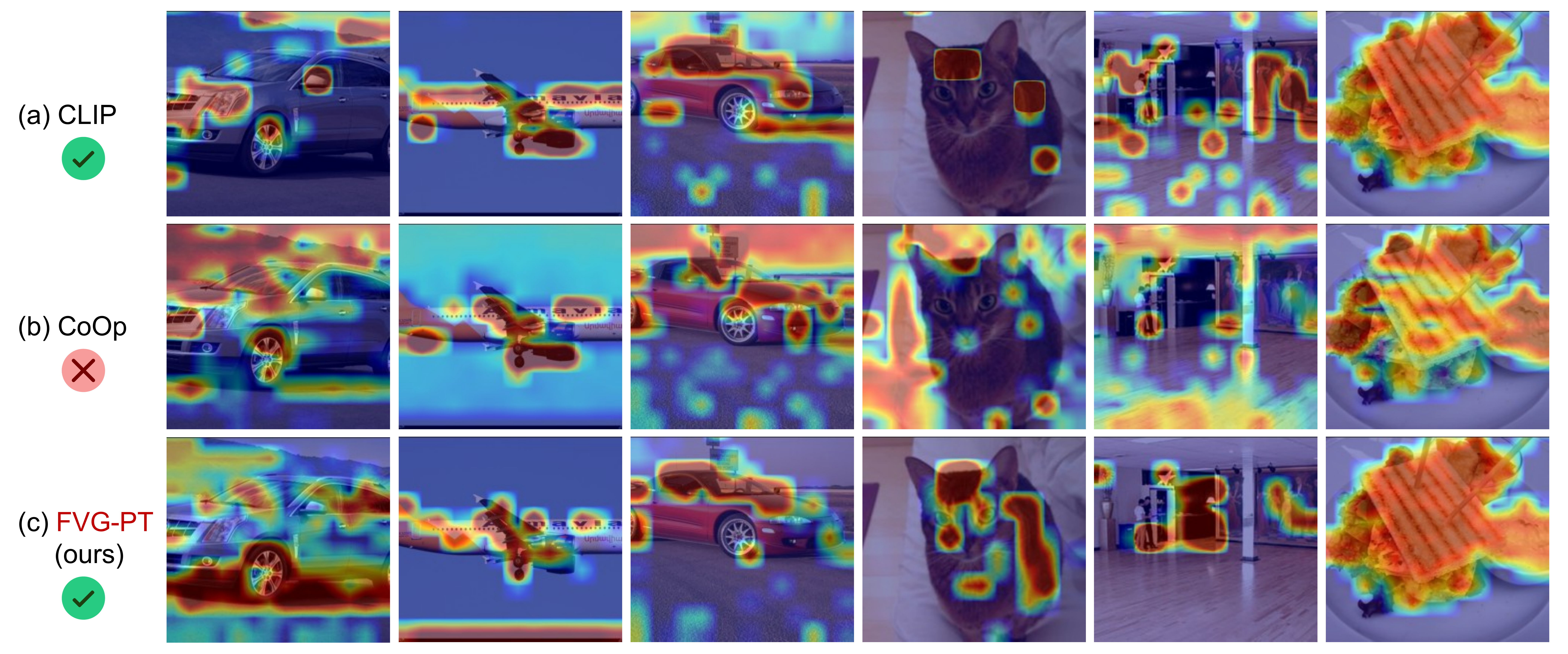}
  \caption{Failure cases of (b) CoOp compared with (a) the original CLIP and (c) our \texttt{FVG-PT}. Attention maps of visual encoders are generated by Grad-CAM \cite{selvaraju2017gradcam}.}
  \label{Fig-S3}
\end{figure}

We quantitatively evaluate how attention shifts in attention affect performance on the target task (base classes) on the entire test set of the datasets. Specifically, we define a \textit{foreground shift index} based on foreground consistency, denoted as the KL divergence between the full-image and foreground prediction distributions $D_{\text{fg}}=\operatorname{KL}\left(\mathbf{p}_{\mathrm{full}} \| \mathbf{p}_{\mathrm{fg}}\right)$. For an input image at inference time, if the discriminative evidence for prompt tuning predictions focuses more on the foreground, then its prediction distribution on the full image should be more consistent with that on the foreground view, leading to a smaller $D_{\text{fg}}$. In contrast, a larger $D_{\text{fg}}$ indicates that the model does not primarily rely on foreground regions for prediction, thereby reflecting an attention shift away from the main object.

As examples, Fig.~\ref{Fig-S1} reports the foreground shift index and base-class performance for 3 progressively refined models, namely the foundation CLIP \cite{radford2021clip}, fine-tuned CoOp \cite{zhou2022coop} built on CLIP, and \texttt{FVG-PT} that further focuses on the foreground based on CoOp. We observe that the base performance increases while the foreground shift index decreases. In other words, for CoOp and \texttt{FVG-PT}, the attention shift with respect to the image foreground is gradually corrected, which in turn leads to higher accuracy than CLIP.

\subsection{More Visualization Results}
In this section, building on Fig.~\ref{Fig-1} in the main text, we present additional Grad-CAM based attention visualizations as case studies to illustrate the failure modes of CLIP and CoOp. Fig.~\ref{Fig-S2} reports examples where CLIP makes incorrect predictions while CoOp succeeds, and Fig.~\ref{Fig-S3} reports cases where CoOp fails but CLIP predicts correctly.

Across these images, the attention maps exhibit a pattern similar to Fig.~\ref{Fig-1}, that is, for misclassified examples, the visual encoder tends to shift attention away from the foreground. This observation motivates \texttt{FVG-PT}, which guides the visual encoder to correct such shifts in attention, and thereby improves prediction accuracy.

\section{More Implementation Details} \label{sec:A}

\subsection{Experimental Setup} \label{sec:A1}

\begin{table}[t]
    \caption{Training settings of backbones on base-to-new tasks.}
    \label{tabS1}
    \centering
    \setlength\tabcolsep{6pt}
    \scalebox{0.8}{
\begin{tabular}{lcccc} 
\toprule
\rowcolor{gray!10} \textbf{Params} & \textbf{CoOp} & \textbf{KgCoOp} & \textbf{PromptSRC} & \textbf{MMRL}  \\ 
\midrule
Depth of text prompt                              & 1             & 1               & 9                  & -              \\
Depth of visual prompt                            & -             & -               & 9                  & -              \\
Text token length                                 & 4             & 4               & 4                  & 5              \\
Visual token length                               & -             & -               & 4                  & 5              \\
Layers that insert learnable tokens                        & 1             & 1               & 1-9                & 6-12           \\
Optimizer                                         & SGD           & SGD             & SGD                & AdamW          \\
\bottomrule
\end{tabular}
}
\end{table}

As a plug-and-play module, \texttt{FVG-PT} is evaluated on 4 prompt tuning backbones \cite{zhou2022coop, yao2023kgcoop, khattak2023promptsrc, guo2025mmrl} that all use a ViT-B/16-based CLIP as the foundation model. It shares exactly the same fine-tuning data and initialization parameters with each backbone to ensure a fair comparison. The basic hyperparameters of the backbones are summarized in Tab.~\ref{tabS1}, and their architectures and initialization procedures are detailed in \textit{Supplementary Material~\ref{sec:A4}}. For the learnable parameters in prompt tuning, CoOp \cite{zhou2022coop} and KgCoOp \cite{yao2023kgcoop} are randomly initialized from a zero-mean Gaussian $X \sim \mathcal{N}(0, 0.02^2)$. PromptSRC \cite{khattak2023promptsrc} is initialized from the tokenized prompt template ``A photo of a [CLASS]'', and MMRL \cite{guo2025mmrl} uses the same Gaussian initialization $X \sim \mathcal{N}(0, 0.02^2)$ for the mid-layer encoder plugins, with the representation space for cross-modal feature interaction set to dimension 512. We use SEEM (\textit{Supplementary Material~\ref{sec:A2}}) to pre-construct offline foreground masks.

During fine-tuning, we set the batch size to $bs=4$ and use $bs=100$ for inference. For MMRL, the weight of the FDC distillation loss (Sec.~\ref{sec3.3} of the main text) is set to $\lambda_{d}=5.0$, while other models use $\lambda_{d}=10.0$. For the cross-dataset transfer experiments, we follow the PromptSRC protocol and fine-tune on ImageNet using all categories with $ep=5$ and $lr=0.0035$. All experiments are conducted on a single NVIDIA V100 GPU with 3 runs on each dataset.

\begin{table}[t]
    \caption{Textual prompts used in SEEM model for generating foreground masks.}
    \label{tabS2}
    \centering
    \setlength\tabcolsep{8pt}
    \scalebox{0.8}{
\begin{tabular}{cl} 
\toprule
\rowcolor{gray!10} \textbf{Dataset} & \textbf{SEEM Textual Prompt}      \\ 
\midrule
ImageNet                                           & The whole [CLASS] in the middle.  \\
Caltech101                                         & A huge [CLASS].                   \\
Food101                                            & The whole [CLASS] in the middle.  \\
StanfordCars                                       & A huge and whole car.             \\
OxfordPets                                         & The pet in the middle.            \\
Flowers102                                         & The flower in the middle.         \\
DTD                                                & The [CLASS].                      \\
EuroSAT                                            & The satellite [CLASS].            \\
FGVCAircraft                                       & The huge airplane in the middle.  \\
SUN397                                             & The [CLASS].                      \\
UCF101                                             & The people with [CLASS].          \\
\bottomrule
\end{tabular}
}
\end{table}
\begin{figure}[t]
  \centering
  \includegraphics[width=\textwidth]{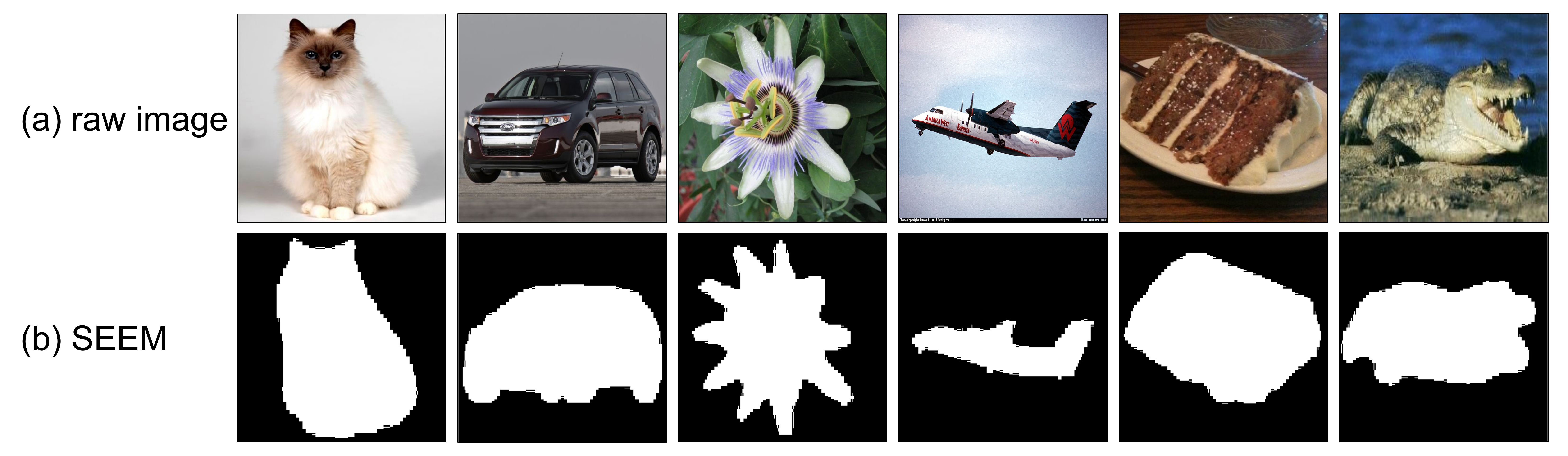}
  \caption{Illustration of the foreground mask generated by SEEM \cite{zou2023seem}, where (a) shows the raw image and (b) presents the corresponding foreground mask visualization.}
  \label{Fig-S4}
\end{figure}

\subsection{Details of Segmentation Model} \label{sec:A2}
In this paper, we use an advanced pretrained SEEM \cite{zou2023seem} semantic segmentation model as an optional approach to obtain pixel-level foreground masks. SEEM (Segment Everything Everywhere All at Once) is a promptable and interactive general segmentation model that accepts an image and a textual prompt as input and supports a variety of segmentation tasks in a unified manner. Specifically, we use a strategy similar to DAPT \cite{zhang2025dapt}, directly using \textit{the simplest, un-optimized textual prompts} as queries to generate foreground views, as summarized in Tab.~\ref{tabS2}. For fine-grained object names that SEEM may find difficult to recognize (e.g., specific car models in StanfordCars or animal breeds in OxfordPets), we use same object name (e.g., ``car'' or ``pet'') across the entire dataset.

It is important to note that, because \texttt{FVG-PT} can adaptively assess foreground quality through FRG, the prompts used for SEEM do not require careful manual design or optimization. During fine-tuning, the data processing speed of SEEM is $\sim190$ FPS, which does not constitute a computational bottleneck.

\subsection{FVG-PT Optimization for Backbones} \label{sec:A4}

In this section, we briefly introduce the frameworks of the 4 backbones used by \texttt{FVG-PT} and explain how \texttt{FVG-PT} integrates seamlessly with them.

\paragraph{\bf CoOp \cite{zhou2022coop}.}  CoOp directly introduces a set of randomly initialized learnable prompt tokens to replace the fixed textual template ``A photo of a [CLASS]'' in the foundation CLIP and is fine-tuned with the cross-entropy loss $\mathcal{L}_{\mathrm{CE}}$. Since CoOp performs image-text alignment directly after the standard ViT encoders, \texttt{FVG-PT} attaches the FDC adapters and the PC branch to the CoOp image-text encoders and then performs foreground-guided fine-tuning on top of the pre-tuned CoOp backbone.

\paragraph{\bf KgCoOp \cite{yao2023kgcoop}.}  KgCoOp uses the same learnable parameters as CoOp, but integrates prior textual prompt templates as additional knowledge and applies a knowledge-guided loss $\mathcal{L}_{\mathrm{KG}}$ based on minimizing Euclidean distance together with the standard cross-entropy loss  $\mathcal{L}_{\mathrm{CE}}$ for fine-tuning. Since this injected knowledge does not alter the original image-text feature embedding pattern of the ViT encoders, \texttt{FVG-PT} inserts its modules into KgCoOp in the same way as in CoOp and performs foreground-guided fine-tuning on the frozen pre-tuned KgCoOp backbone.

\paragraph{\bf PromptSRC \cite{khattak2023promptsrc}.}  PromptSRC is a cross-modal prompting model that adopts the IVLP \cite{rasheed2023ivlp} setting, where modality-specific prompt vectors are independently injected into the visual and textual branches, and inserted into both the model input and the first $K$ layers of the 12 Transformer blocks in the ViT encoder. During fine-tuning, in addition to the standard cross-entropy loss $\mathcal{L}_{\mathrm{CE}}$, PromptSRC introduces 3 consistency-based losses to alleviate the BNT problem \cite{li2025dpc}, which are consistency between prompts and their corresponding modality features ($\mathcal{L}_{\mathrm{SCL-image}}$ and $\mathcal{L}_{\mathrm{SCL-text}}$) and consistency between the logits after cross-modal interaction ($\mathcal{L}_{\mathrm{SCL-logits}}$). When initializing \texttt{FVG-PT}, we load and freeze all prompt vectors from the pre-tuned PromptSRC backbone, and then append branch-specific FDC adapters and PC branches after the last Transformer block of the corresponding encoders in the visual and textual branches.

\begin{table}[t]
    \caption{Error bar analysis of \texttt{FVG-PT} on 4 backbone models.}
    \label{tabS3}
    \centering
    \setlength\tabcolsep{2pt}
    \scalebox{0.8}{
\begin{tabular}{c|cccccc} 
\toprule
\rowcolor{gray!10} \textbf{Method} & \textbf{Avg.}       & \textbf{ImageNet} & \textbf{Caltech101} & \textbf{Food101}      & \textbf{StanfordCars} & \textbf{OxfordPets}  \\ 
\midrule
CoOp                                           & 75.09               & 72.27             & 96.20               & 90.81                 & 69.43                 & 96.07                \\
\textbf{+FVG-PT}                               & 77.16 (±0.20)       & 73.28 (±0.02)     & 96.42 (±0.02)       & 90.57 (±0.01)         & 74.66 (±0.22)         & 95.81 (±0.14)        \\ 
\midrule
KgCoOp                                         & 76.88               & 73.10             & 96.16               & 91.12                 & 72.44                 & 96.25                \\
\textbf{+FVG-PT}                               & 78.10 (±0.15)       & 73.55 (±0.04)     & 96.48 (±0.03)       & 90.84 (±0.03)         & 75.28 (±0.17)         & 96.28 (±0.04)        \\ 
\midrule
PromptSRC                                      & 77.24               & 73.66             & 96.12               & 91.27                 & 73.73                 & 96.47                \\
\textbf{+FVG-PT}                               & 77.89 (±0.12)       & 73.82 (±0.01)     & 96.32 (±0.02)       & 91.25 (±0.03)         & 75.03 (±0.07)         & 96.28 (±0.03)        \\ 
\midrule
MMRL                                           & 80.10               & 74.48             & 96.43               & 90.75                 & 77.83                 & 96.31                \\
\textbf{+FVG-PT}                               & 80.75 (±0.11)       & 74.48 (±0.02)     & 96.49 (±0.03)       & 90.64 (±0.06)         & 79.00 (±0.14)         & 96.44 (±0.05)        \\ 
\hline\hline
\rowcolor{gray!10} \textbf{Method} & \textbf{Flowers102} & \textbf{DTD}      & \textbf{EuroSAT}    & \textbf{FGVCAircraft} & \textbf{SUN397}       & \textbf{UCF101}      \\ 
\midrule
CoOp                                           & 82.18               & 60.01             & 60.41               & 33.14                 & 76.89                 & 80.25                \\
\textbf{+FVG-PT}                               & 85.04 (±0.19)       & 64.83 (±0.80)     & 70.67 (±0.12)       & 37.31 (±0.32)         & 79.35 (±0.17)         & 82.20 (±0.18)        \\ 
\midrule
KgCoOp                                         & 82.80               & 67.65             & 69.76               & 34.94                 & 78.47                 & 79.71                \\
\textbf{+FVG-PT}                               & 84.32 (±0.26)       & 68.75 (±0.65)     & 73.94 (±0.08)       & 36.32 (±0.19)         & 79.35 (±0.13)         & 81.06 (±0.11)        \\ 
\midrule
PromptSRC                                      & 83.07               & 69.23             & 67.19               & 35.67                 & 79.84                 & 80.54                \\
\textbf{+FVG-PT}                               & 84.10 (±0.16)       & 71.41 (±0.41)     & 68.07 (±0.19)       & 36.54 (±0.15)         & 80.26 (±0.07)         & 81.36 (±0.23)        \\ 
\midrule
MMRL                                           & 86.47               & 74.03             & 76.85               & 39.56                 & 80.91                 & 83.79                \\
\textbf{+FVG-PT}                               & 86.90 (±0.25)       & 75.42 (±0.34)     & 77.89 (±0.02)       & 41.81 (±0.17)         & 81.02 (±0.02)         & 84.64 (±0.09)        \\
\bottomrule
\end{tabular}
}
\end{table}

\paragraph{\bf MMRL \cite{guo2025mmrl}.}  MMRL adapts to downstream tasks by introducing learnable mid-layer plugins into the encoders of the image and text branches. Unlike the above models, MMRL uses a fixed prompt template at the input, and constructs a shared cross-modal representation space to learn task-specific features, which are projected into the last $K$ Transformer blocks of the two branches. To integrate with this type of model, \texttt{FVG-PT} does not modify the mid-layer feature transformation and embedding process. Instead, it takes the final-block features from the MMRL-optimized image and text encoders and feeds them into the respective FDC adapters and PC branches.

It is also worth noting that MMRL applies different feature processing pipelines to the base and new branches. Since the PC branch in \texttt{FVG-PT} for new classes is fully decoupled from the base branch, we do not alter the original MMRL branches. Rather, for each branch, we perform additional foreground-guided fine-tuning on the logits after image-text interaction.

\begin{table}[t]
    \caption{Comparison with additional prompt tuning baselines.}
    \label{tabS4}
    \centering
    \setlength\tabcolsep{5pt}
    \scalebox{0.8}{
\begin{tabular}{cc|ccc} 
\toprule
\rowcolor{gray!10} {\cellcolor{gray!10}}                                  & {\cellcolor{gray!10}}                                 & \multicolumn{3}{c}{\textbf{Avg. over 11}}                                                                        \\
\rowcolor{gray!10} \multirow{-2}{*}{{\cellcolor{gray!10}}\textbf{Method}} & \multirow{-2}{*}{{\cellcolor{gray!10}}\textbf{Model}} & Base                                & New                                 & HM                                    \\ 
\midrule
zero-shot                                                                                               & CLIP \cite{radford2021clip}                                                                 & 69.34                               & 74.22                               & 71.70                                \\ 
\midrule
\multirow{3}{*}{\begin{tabular}[c]{@{}c@{}}prompt\\optimization\end{tabular}}                           & CoCoOp \cite{zhou2022cocoop}                                                              & 80.47                               & 71.69                               & 75.83                                \\
                                                                                                        & ProDA \cite{lu2022proda}                                                                & 81.56                               & 72.30                               & 76.65                                \\
                                                                                                        & MaPLe \cite{khattak2023maple}                                                               & 82.28                               & 75.14                               & 78.55                                \\ 
\midrule
\multirow{2}{*}{\begin{tabular}[c]{@{}c@{}}knowledge\\integration\end{tabular}}                         & KAPT \cite{kan2023kapt}                                                                & 78.41                               & 70.45                               & 74.22                                \\
                                                                                                        & KDPL \cite{mistretta2024kdpl}                                                                & 77.11                               & 71.61                               & 74.26                                \\ 
\midrule
\multirow{4}{*}{\begin{tabular}[c]{@{}c@{}}mid-layer\\plugins\end{tabular}}                             
& CLIP-Adapter \cite{gao2024clipadapter}    & 80.83     & 72.93     & 76.67 \\
& TCP \cite{yao2024tcp}                                                                 & 84.13                               & 75.36                               & 79.51                                \\
                                                                                                        & MMA \cite{seputis2024mma}                                                                 & 83.20                               & 76.80                               & 79.87                                \\
                                                                                                        & {\cellcolor{cyan!10}}FVG-PT+MMRL \cite{guo2025mmrl}                            & {\cellcolor{cyan!10}}86.04 & {\cellcolor{cyan!10}}76.06 & {\cellcolor{cyan!10}}80.75  \\
\bottomrule
\end{tabular}
}
\end{table}

\section{More Experimental Results} \label{sec:B}

\subsection{Error Bar Analysis} \label{sec:B1}

To assess the robustness of \texttt{FVG-PT} under different random seeds, we fine-tune \texttt{FVG-PT} with $seed=1, 2, 3$ for splitting the data, and run 3 trials accordingly. To avoid data leakage, the same seeds are also used for the backbone models so that \texttt{FVG-PT} and the corresponding backbones always see exactly the same sampled data in each run. The results of base-to-new tasks are reported in Tab.~\ref{tabS3}, including both the mean performance and the standard deviation (SD).

In most datasets, \texttt{FVG-PT} reports relatively small standard deviations, indicating stable behavior under different fine-tuning conditions. For some fine-grained datasets (e.g., Flowers102, DTD, and FGVCAircraft), the standard deviation is slightly larger. A possible explanation is that the quality of the foreground views is less stable on fine-grained categories, so different splits provide different amounts of reliable foreground information for \texttt{FVG-PT} (which interacts with the FRG-based quality control) and leads to performance fluctuation. Even in these cases, the overall performance of \texttt{FVG-PT} remains consistently higher than that of the backbone models.

\subsection{Compare with More Baselines} \label{sec:B2}
To further highlight the performance advantages of our \texttt{FVG-PT}, Tab.~\ref{tabS4} reports the average HM performance on base-to-new tasks for zero-shot CLIP and mainstream prompt tuning models. These models include CoCoOp \cite{zhou2022cocoop}, ProDA \cite{lu2022proda}, MaPLe \cite{khattak2023maple}, KAPT \cite{kan2023kapt}, KDPL \cite{mistretta2024kdpl}, CLIP-Adapter \cite{gao2024clipadapter}, TCP \cite{yao2024tcp}, and MMA \cite{seputis2024mma}. We observe that the combination of \texttt{FVG-PT} and MMRL achieves higher performance than all of the above models, thereby demonstrating its superiority.

\begin{table}[t]
    \caption{Ablation study of losses in \texttt{FVG-PT} on base-to-new tasks.}
    \label{tabS5}
    \centering
    \setlength\tabcolsep{6pt}
    \scalebox{0.8}{
\begin{tabular}{cccccc|ccc|c} 
\toprule
\rowcolor{gray!10} {\cellcolor{gray!10}}                   & FRG                                                               & \multicolumn{2}{c}{FDC}                                                                                                               & \multicolumn{2}{c|}{PC}                                                                                                            & \multicolumn{3}{c|}{Average over 11} & {\cellcolor{gray!10}}                         \\
\rowcolor{gray!10} \multirow{-2}{*}{{\cellcolor{gray!10}}} & $\mathcal{L}_{\text {FRG}}$ & $\mathcal{L}_{\text{CE}}$ & $\mathcal{L}_{\text {dist}}$ & $\mathcal{L}_{\text {CE}}$ & $\mathcal{L}_{\text {KL}}$ & Base  & New   & HM                            & \multirow{-2}{*}{{\cellcolor{gray!10}}\textbf{$\Delta$}}  \\ 
\midrule
                                                                                         & \ding{55}                                                            & \ding{55}                                                           & \ding{55}                                                             & \ding{55}                                                           & \ding{55}                                                           & 80.94 & 70.03 & 75.09                & (baseline)                                                   \\
(6)                                                                                      & \ding{51}                                                            & \ding{55}                                                           & \ding{51}                                                             & \ding{51}                                                           & \ding{51}                                                           & 81.01 & 73.62 & 77.14                & \textcolor{V}{+ 2.05}                                                       \\
(7)                                                                                      & \ding{51}                                                            & \ding{51}                                                           & \ding{55}                                                             & \ding{51}                                                           & \ding{51}                                                           & 81.56 & 73.62 & 77.38                & \textcolor{V}{+ 2.29}                                                       \\
(8)                                                                                      & \ding{51}                                                            & \ding{51}                                                           & \ding{51}                                                             & \ding{55}                                                           & \ding{51}                                                           & 82.40 & 72.58 & 77.18                & \textcolor{V}{+ 2.09}                                                       \\
(9)                                                                                      & \ding{51}                                                            & \ding{51}                                                           & \ding{51}                                                             & \ding{51}                                                           & \ding{55}                                                           & 82.40 & 71.11 & 76.33                & \textcolor{V}{+ 1.24}                                                       \\
                                                                                         & \ding{51}                                                            & \ding{51}                                                           & \ding{51}                                                             & \ding{51}                                                           & \ding{51}                                                           & 82.40 & 73.62 & 77.76                & (full)                                                       \\
\bottomrule
\end{tabular}
}
\end{table}

\subsection{Detailed Ablation Study} \label{sec:B3}

\paragraph{\bf Losses of \texttt{FVG-PT}.}  Building on Tab.~\ref{tab3} in the main text, this section presents a more fine-grained loss-level ablation to validate the effectiveness of each loss in \texttt{FVG-PT}. Specifically, we study 5 losses used during fine-tuning: $\mathcal{L}_{\text{FRG}}$ in the FRG module (Sec.~\ref{sec3.2} of the main text) for training the adaptive foreground trust score, $\mathcal{L}_{\text{dist}}$ and $\mathcal{L}_{\text{CE}}$ in the FDC module (Sec.~\ref{sec3.3} of the main text) for guiding attention toward the foreground view, and $\mathcal{L}_{\text{CE}}$ and $\mathcal{L}_{\text{KL}}$ in the PC module of the new branch (Sec.~\ref{sec3.4} of the main text) for BRG to learn the weight assigned to the CLIP prior.

Tab.~\ref{tabS5} shows that removing any one of these losses prevents the model from reaching its best performance, which indicates that all five losses contribute positively to the fine-tuning of \texttt{FVG-PT}. Among them, the KL loss in PC module has the most pronounced impact on new-class accuracy, suggesting that aligning the weighted distribution $\mathbf{p}_{\text{PC}}$ with the CLIP prior $\mathbf{p}_{\text{CLIP}}$ is crucial for strong generalization on the new branch.

\begin{table}[t]
    \caption{Ablation study of the branches of FDC adapters (Sec.~\ref{sec3.3}) in \texttt{FVG-PT} on base-to-new tasks.}
    \label{tabS8}
    \centering
    \setlength\tabcolsep{6pt}
    \scalebox{0.8}{
\begin{tabular}{cc|ccc|c} 
\toprule
\rowcolor{gray!10} {\cellcolor{gray!10}}                                  & {\cellcolor{gray!10}}                                  & \multicolumn{3}{c|}{\textbf{Average over 11}} & {\cellcolor{gray!10}}                                  \\
\rowcolor{gray!10} \multirow{-2}{*}{{\cellcolor{gray!10}}\textbf{Method}} & \multirow{-2}{*}{{\cellcolor{gray!10}}\textbf{Branch}} & Base  & New   & HM                            & \multirow{-2}{*}{{\cellcolor{gray!10}}\textbf{$\Delta$}}  \\ 
\midrule
CoOp                                                                                                    & -                                                                     & 80.94 & 70.03 & 75.09                         & (baseline)                                                            \\ 
\midrule
\multirow{2}{*}{\textbf{+FVG-PT}}                                                                       & visual + textual                                                            & 82.40 & 73.62 & 77.76                         & \textcolor{V}{+2.67}                                                                 \\
                                                                                                        & visual only                                                            & 81.80 & 73.62 & 77.49                         & \textcolor{V}{+2.40}                                                                 \\
\bottomrule
\end{tabular}
}
\end{table}

\begin{table}[t]
    \caption{Ablation study of the ViT encoders of foundation CLIP used in CoOp \cite{zhou2022coop} and \texttt{FVG-PT} on base-to-new tasks.}
    \label{tabS9}
    \centering
    \setlength\tabcolsep{6pt}
    \scalebox{0.8}{
\begin{tabular}{cc|ccc|c} 
\toprule
\rowcolor{gray!10} {\cellcolor{gray!10}}                                                                                                                               & {\cellcolor{gray!10}}                                  & \multicolumn{3}{c|}{\textbf{Average over 11}}                                                                         & {\cellcolor{gray!10}}                                  \\
\rowcolor{gray!10} \multirow{-2}{*}{{\cellcolor{gray!10}}\begin{tabular}[c]{@{}>{\cellcolor{gray!10}}c@{}}\textbf{Foundation}\\\textbf{CLIP}\end{tabular}} & \multirow{-2}{*}{{\cellcolor{gray!10}}\textbf{Method}} & Base                                  & New                                   & HM                                    & \multirow{-2}{*}{{\cellcolor{gray!10}}\textbf{$\Delta$}}  \\ 
\midrule
\multirow{2}{*}{ViT-B/16}                                                                                                                                                                      & CoOp                                                               & 80.94                                 & 70.03                                 & 75.09                                 &                                                                    \\
                                                                                                                                                                                               & {\cellcolor{cyan!10}}\textbf{+FVG-PT}                   & {\cellcolor{cyan!10}}82.40 & {\cellcolor{cyan!10}}73.62 & {\cellcolor{cyan!10}}77.76 & {\cellcolor{cyan!10}}\textcolor{V}{+2.67}                              \\ 
\midrule
\multirow{2}{*}{ViT-L/14}                                                                                                                                                                      & CoOp                                                               & 83.40                                 & 75.05                                 & 79.01                                 &                                                                    \\
                                                                                                                                                                                               & {\cellcolor{cyan!10}}\textbf{+FVG-PT}                   & {\cellcolor{cyan!10}}83.98 & {\cellcolor{cyan!10}}80.39 & {\cellcolor{cyan!10}}82.15 & {\cellcolor{cyan!10}}\textcolor{V}{+3.14}                              \\
\bottomrule

\end{tabular}
}
\end{table}

\begin{table}[t]
    \caption{Failure case study of \texttt{FVG-PT} on EuroSAT dataset.}
    \label{tabS7}
    \centering
    \setlength\tabcolsep{6pt}
    \scalebox{0.8}{
\begin{tabular}{cc|ccc|c} 
\toprule
\rowcolor{gray!10} {\cellcolor{gray!10}}                                  & {\cellcolor{gray!10}}                                 & \multicolumn{3}{c|}{\textbf{EuroSAT}} & {\cellcolor{gray!10}}                                  \\
\rowcolor{gray!10} \multirow{-2}{*}{{\cellcolor{gray!10}}\textbf{Method}} & \multirow{-2}{*}{{\cellcolor{gray!10}}\textbf{epoch}} & Base  & New   & HM                     & \multirow{-2}{*}{{\cellcolor{gray!10}}\textbf{$\Delta$}}  \\ 
\midrule
CoOp                                                                                                    & 10                                                                   & 88.43 & 45.87 & 60.41                 &                                                                       \\ 
\midrule
\multirow{3}{*}{\textbf{+FVG-PT}}                                                                       & 10                                                                   & \textcolor{X}{72.62} & 58.15 & 64.58                 & \textcolor{V}{+4.18}                                                                 \\
                                                                                                        & 15                                                                   & 86.98 & 58.21 & 69.74                 & \textcolor{V}{+9.34}                                                                 \\
                                                                                                        & 20                                                                   & 89.76 & 58.28 & 70.67                 & \textcolor{V}{+10.27}                                                                \\
\bottomrule
\end{tabular}
}
\end{table}

\begin{figure}[t]
  \centering
  \includegraphics[width=0.55\textwidth]{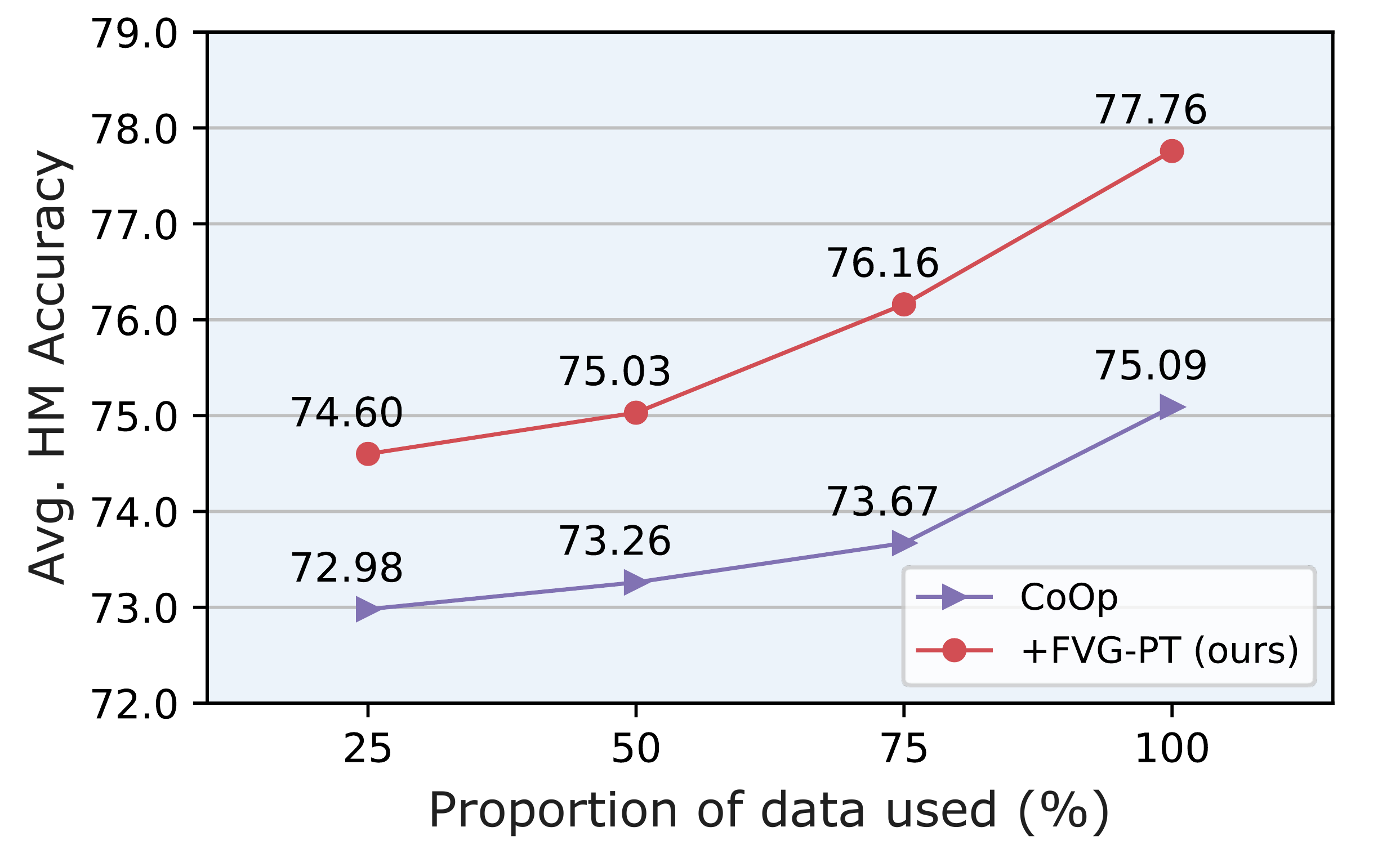}
  \caption{Data efficiency analysis using different proportions of the 11 datasets as the training set on base-to-new tasks.}
  \label{Fig-S5}
\end{figure}

\paragraph{\bf Adapters in Image and Text Branches.}  In Sec.~\ref{sec3.3} of the main text, \texttt{FVG-PT} introduces separate adapters with non-shared parameters into both the visual and textual branches of the FDC module to learn a foreground-oriented feature re-projection. To verify the necessity of this design, we report an ablation on adapters in Tab.~\ref{tabS8}. Specifically, since feature re-projection in the visual branch is necessary, we construct a variant that removes the adapter from the text branch while keeping all other fine-tuning settings unchanged. We observe that \texttt{FVG-PT} with a visual-only adapter shows lower base performance than the dual-branch adapter configuration. This result indicates that inserting adapters into both the visual and textual branches helps \texttt{FVG-PT} maintain cross-modal alignment during fine-tuning.

\paragraph{\bf Scaling of ViT Encoders in Foundation CLIP.}  In this section, we adopt more powerful ViT-L/14 encoders in CLIP as the foundation model for both the CoOp backbone and \texttt{FVG-PT} to assess the performance gain of \texttt{FVG-PT} on a scaled CLIP. Results in Tab.~\ref{tabS9} show that \texttt{FVG-PT} achieves clear improvements over the CoOp backbone under both encoder settings, with a larger boost on new classes when using ViT-L/14. This observation suggests that a stronger CLIP prior allows \texttt{FVG-PT} to unlock better generalization on new classes.

\subsection{Data Efficiency Analysis} \label{sec:B5}
In this section, we examine how the performance of \texttt{FVG-PT} and the CoOp backbone varies under different proportions of training data (25\%, 50\%, 75\%, 100\%). To avoid data leakage, \texttt{FVG-PT} is always fine-tuned on top of a CoOp backbone trained with exactly the same subset of data. Fig.~\ref{Fig-S5} shows that \texttt{FVG-PT} already matches the HM performance of CoOp with full data when only 50\% of the training samples are used, which indicates that \texttt{FVG-PT} achieves better data efficiency and requires fewer image-text pairs to reach a similar accuracy level.

\section{Failure Case Study} \label{sec:C}

\paragraph{\bf Performance Degradation of \texttt{FVG-PT}+CoOp on EuroSAT.}   In the experiments of \texttt{FVG-PT}+CoOp, we observe that using the default configuration in Sec.~\ref{sec3.1} of the main text leads to a severe drop in base performance on EuroSAT dataset, from 88.43 to 72.62 ($-15.81$ points). We attribute this behavior to the very limited data scale on EuroSAT, where the base set contains only 5 classes, and the 16-shot setting provides only 80 image-text pairs for prompt tuning. This scale is much smaller than the amount of learnable parameters in the FDC adapters (0.06M, Sec.~\ref{sec3.3} of the main text), which causes underfitting during fine-tuning and degrades the alignment of the re-projected features. As supporting evidence, the new-class branch of \texttt{FVG-PT}, which has only 0.19K learnable parameters, achieves a clear improvement on new classes under the same setting.

To mitigate this degradation, Tab.~\ref{tabS7} evaluates a simple solution that increases the number of training epochs from 10 to 15 and 20. The results show that with $ep = 20$, \texttt{FVG-PT} achieves higher base performance than CoOp and also better new generalization than the $ep = 10$  setting, resulting in an HM improvement of 10.27 points over the CoOp backbone. Therefore, in the main experiments (Tab.~\ref{tab1} and Tab.~\ref{tab2} of the main text), we set $ep = 20$ for EuroSAT on the CoOp backbone and \texttt{FVG-PT} for a better performance. Since EuroSAT contains very few samples, increasing the number of epochs does not noticeably increase the computational cost.

%% file: main.bib
@String(AAAI  = {AAAI})

@String(ICME  = {Int. Conf. Multimedia and Expo})

@String(ICME  =	{ICME})

@inproceedings{radford2021clip,
  title={Learning transferable visual models from natural language supervision},
  author={Radford, Alec and Kim, Jong Wook and Hallacy, Chris and Ramesh, Aditya and Goh, Gabriel and Agarwal, Sandhini and Sastry, Girish and Askell, Amanda and Mishkin, Pamela and Clark, Jack and others},
  booktitle={International conference on machine learning},
  pages={8748--8763},
  year={2021},
  organization={PMLR}
}

@article{zhou2022coop,
  title={Learning to prompt for vision-language models},
  author={Zhou, Kaiyang and Yang, Jingkang and Loy, Chen Change and Liu, Ziwei},
  journal={International Journal of Computer Vision},
  volume={130},
  number={9},
  pages={2337--2348},
  year={2022},
  publisher={Springer}
}

@inproceedings{zhou2022cocoop,
  title={Conditional prompt learning for vision-language models},
  author={Zhou, Kaiyang and Yang, Jingkang and Loy, Chen Change and Liu, Ziwei},
  booktitle={Proceedings of the IEEE/CVF conference on computer vision and pattern recognition},
  pages={16816--16825},
  year={2022}
}

@inproceedings{zhang2024dept,
  title={Dept: Decoupled prompt tuning},
  author={Zhang, Ji and Wu, Shihan and Gao, Lianli and Shen, Heng Tao and Song, Jingkuan},
  booktitle={Proceedings of the IEEE/CVF Conference on Computer Vision and Pattern Recognition},
  pages={12924--12933},
  year={2024}
}

@inproceedings{yao2023kgcoop,
  title={Visual-language prompt tuning with knowledge-guided context optimization},
  author={Yao, Hantao and Zhang, Rui and Xu, Changsheng},
  booktitle={Proceedings of the IEEE/CVF conference on computer vision and pattern recognition},
  pages={6757--6767},
  year={2023}
}

@inproceedings{khattak2023promptsrc,
  title={Self-regulating prompts: Foundational model adaptation without forgetting},
  author={Khattak, Muhammad Uzair and Wasim, Syed Talal and Naseer, Muzammal and Khan, Salman and Yang, Ming-Hsuan and Khan, Fahad Shahbaz},
  booktitle={Proceedings of the IEEE/CVF International Conference on Computer Vision},
  pages={15190--15200},
  year={2023}
}

@inproceedings{yao2024tcp,
  title={TCP: Textual-based Class-aware Prompt tuning for Visual-Language Model},
  author={Yao, Hantao and Zhang, Rui and Xu, Changsheng},
  booktitle={Proceedings of the IEEE/CVF Conference on Computer Vision and Pattern Recognition},
  pages={23438--23448},
  year={2024}
}

@article{gao2024clipadapter,
  title={Clip-adapter: Better vision-language models with feature adapters},
  author={Gao, Peng and Geng, Shijie and Zhang, Renrui and Ma, Teli and Fang, Rongyao and Zhang, Yongfeng and Li, Hongsheng and Qiao, Yu},
  journal={International Journal of Computer Vision},
  volume={132},
  number={2},
  pages={581--595},
  year={2024},
  publisher={Springer}
}

@article{roy2023coprompt,
  title={Consistency-guided prompt learning for vision-language models},
  author={Roy, Shuvendu and Etemad, Ali},
  journal={arXiv preprint arXiv:2306.01195},
  year={2023}
}

@inproceedings{li2024promptkd,
  title={Promptkd: Unsupervised prompt distillation for vision-language models},
  author={Li, Zheng and Li, Xiang and Fu, Xinyi and Zhang, Xin and Wang, Weiqiang and Chen, Shuo and Yang, Jian},
  booktitle={Proceedings of the IEEE/CVF Conference on Computer Vision and Pattern Recognition},
  pages={26617--26626},
  year={2024}
}

@inproceedings{khattak2023maple,
  title={Maple: Multi-modal prompt learning},
  author={Khattak, Muhammad Uzair and Rasheed, Hanoona and Maaz, Muhammad and Khan, Salman and Khan, Fahad Shahbaz},
  booktitle={Proceedings of the IEEE/CVF Conference on Computer Vision and Pattern Recognition},
  pages={19113--19122},
  year={2023}
}

@article{lu2019vilbert,
  title={Vilbert: Pretraining task-agnostic visiolinguistic representations for vision-and-language tasks},
  author={Lu, Jiasen and Batra, Dhruv and Parikh, Devi and Lee, Stefan},
  journal={Advances in neural information processing systems},
  volume={32},
  year={2019}
}

@inproceedings{jia2021align,
  title={Scaling up visual and vision-language representation learning with noisy text supervision},
  author={Jia, Chao and Yang, Yinfei and Xia, Ye and Chen, Yi-Ting and Parekh, Zarana and Pham, Hieu and Le, Quoc and Sung, Yun-Hsuan and Li, Zhen and Duerig, Tom},
  booktitle={International conference on machine learning},
  pages={4904--4916},
  year={2021},
  organization={PMLR}
}

@inproceedings{li2023blip,
  title={Blip-2: Bootstrapping language-image pre-training with frozen image encoders and large language models},
  author={Li, Junnan and Li, Dongxu and Savarese, Silvio and Hoi, Steven},
  booktitle={International conference on machine learning},
  pages={19730--19742},
  year={2023},
  organization={PMLR}
}

@article{han2024peft,
  title={Parameter-efficient fine-tuning for large models: A comprehensive survey},
  author={Han, Zeyu and Gao, Chao and Liu, Jinyang and Zhang, Sai Qian and others},
  journal={arXiv preprint arXiv:2403.14608},
  year={2024}
}

@article{li2021prefix,
  title={Prefix-tuning: Optimizing continuous prompts for generation},
  author={Li, Xiang Lisa and Liang, Percy},
  journal={arXiv preprint arXiv:2101.00190},
  year={2021}
}

@article{wei2022chain,
  title={Chain-of-thought prompting elicits reasoning in large language models},
  author={Wei, Jason and Wang, Xuezhi and Schuurmans, Dale and Bosma, Maarten and Xia, Fei and Chi, Ed and Le, Quoc V and Zhou, Denny and others},
  journal={Advances in neural information processing systems},
  volume={35},
  pages={24824--24837},
  year={2022}
}

@inproceedings{zhu2023prograd,
  title={Prompt-aligned gradient for prompt tuning},
  author={Zhu, Beier and Niu, Yulei and Han, Yucheng and Wu, Yue and Zhang, Hanwang},
  booktitle={Proceedings of the IEEE/CVF International Conference on Computer Vision},
  pages={15659--15669},
  year={2023}
}

@inproceedings{tian2024argue,
  title={ArGue: Attribute-Guided Prompt Tuning for Vision-Language Models},
  author={Tian, Xinyu and Zou, Shu and Yang, Zhaoyuan and Zhang, Jing},
  booktitle={Proceedings of the IEEE/CVF Conference on Computer Vision and Pattern Recognition},
  pages={28578--28587},
  year={2024}
}

@inproceedings{jia2022vpt,
  title={Visual prompt tuning},
  author={Jia, Menglin and Tang, Luming and Chen, Bor-Chun and Cardie, Claire and Belongie, Serge and Hariharan, Bharath and Lim, Ser-Nam},
  booktitle={European Conference on Computer Vision},
  pages={709--727},
  year={2022},
  organization={Springer}
}

@inproceedings{pei2024sa2vp,
  title={SA$^2$VP: Spatially Aligned-and-Adapted Visual Prompt},
  author={Pei, Wenjie and Xia, Tongqi and Chen, Fanglin and Li, Jinsong and Tian, Jiandong and Lu, Guangming},
  booktitle={Proceedings of the AAAI Conference on Artificial Intelligence},
  pages={4450--4458},
  year={2024}
}

@article{zang2022upt,
  title={Unified vision and language prompt learning},
  author={Zang, Yuhang and Li, Wei and Zhou, Kaiyang and Huang, Chen and Loy, Chen Change},
  journal={arXiv preprint arXiv:2210.07225},
  year={2022}
}

@article{seputis2024mma,
  title={Multi-Modal Adapter for Vision-Language Models},
  author={Seputis, Dominykas and Mihailov, Serghei and Chatterjee, Soham and Xiao, Zehao},
  journal={arXiv preprint arXiv:2409.02958},
  year={2024}
}

@inproceedings{wang2024hpt,
  title={Learning Hierarchical Prompt with Structured Linguistic Knowledge for Vision-Language Models},
  author={Wang, Yubin and Jiang, Xinyang and Cheng, De and Li, Dongsheng and Zhao, Cairong},
  booktitle={Proceedings of the AAAI Conference on Artificial Intelligence},
  pages={5749--5757},
  year={2024}
}

@article{mistretta2024kdpl,
  title={Improving Zero-shot Generalization of Learned Prompts via Unsupervised Knowledge Distillation},
  author={Mistretta, Marco and Baldrati, Alberto and Bertini, Marco and Bagdanov, Andrew D},
  journal={arXiv preprint arXiv:2407.03056},
  year={2024}
}

@article{dosovitskiy2020vit,
  title={An image is worth 16x16 words: Transformers for image recognition at scale},
  author={Dosovitskiy, Alexey},
  journal={arXiv preprint arXiv:2010.11929},
  year={2020}
}

@inproceedings{deng2009imagenet,
  title={Imagenet: A large-scale hierarchical image database},
  author={Deng, Jia and Dong, Wei and Socher, Richard and Li, Li-Jia and Li, Kai and Fei-Fei, Li},
  booktitle={2009 IEEE conference on computer vision and pattern recognition},
  pages={248--255},
  year={2009},
  organization={Ieee}
}

@inproceedings{fei2004caltech,
  title={Learning generative visual models from few training examples: An incremental bayesian approach tested on 101 object categories},
  author={Fei-Fei, Li and Fergus, Rob and Perona, Pietro},
  booktitle={2004 conference on computer vision and pattern recognition workshop},
  pages={178--178},
  year={2004},
  organization={IEEE}
}

@inproceedings{parkhi2012pets,
  title={Cats and dogs},
  author={Parkhi, Omkar M and Vedaldi, Andrea and Zisserman, Andrew and Jawahar, CV},
  booktitle={2012 IEEE conference on computer vision and pattern recognition},
  pages={3498--3505},
  year={2012},
  organization={IEEE}
}

@inproceedings{krause2013cars,
  title={3d object representations for fine-grained categorization},
  author={Krause, Jonathan and Stark, Michael and Deng, Jia and Fei-Fei, Li},
  booktitle={Proceedings of the IEEE international conference on computer vision workshops},
  pages={554--561},
  year={2013}
}

@inproceedings{nilsback2008flowers,
  title={Automated flower classification over a large number of classes},
  author={Nilsback, Maria-Elena and Zisserman, Andrew},
  booktitle={2008 Sixth Indian conference on computer vision, graphics \& image processing},
  pages={722--729},
  year={2008},
  organization={IEEE}
}

@inproceedings{bossard2014food,
  title={Food-101--mining discriminative components with random forests},
  author={Bossard, Lukas and Guillaumin, Matthieu and Van Gool, Luc},
  booktitle={Computer vision--ECCV 2014: 13th European conference, zurich, Switzerland, September 6-12, 2014, proceedings, part VI 13},
  pages={446--461},
  year={2014},
  organization={Springer}
}

@article{maji2013aircraft,
  title={Fine-grained visual classification of aircraft},
  author={Maji, Subhransu and Rahtu, Esa and Kannala, Juho and Blaschko, Matthew and Vedaldi, Andrea},
  journal={arXiv preprint arXiv:1306.5151},
  year={2013}
}

@inproceedings{xiao2010sun,
  title={Sun database: Large-scale scene recognition from abbey to zoo},
  author={Xiao, Jianxiong and Hays, James and Ehinger, Krista A and Oliva, Aude and Torralba, Antonio},
  booktitle={2010 IEEE computer society conference on computer vision and pattern recognition},
  pages={3485--3492},
  year={2010},
  organization={IEEE}
}

@inproceedings{cimpoi2014dtd,
  title={Describing textures in the wild},
  author={Cimpoi, Mircea and Maji, Subhransu and Kokkinos, Iasonas and Mohamed, Sammy and Vedaldi, Andrea},
  booktitle={Proceedings of the IEEE conference on computer vision and pattern recognition},
  pages={3606--3613},
  year={2014}
}

@article{helber2019eurosat,
  title={Eurosat: A novel dataset and deep learning benchmark for land use and land cover classification},
  author={Helber, Patrick and Bischke, Benjamin and Dengel, Andreas and Borth, Damian},
  journal={IEEE Journal of Selected Topics in Applied Earth Observations and Remote Sensing},
  volume={12},
  number={7},
  pages={2217--2226},
  year={2019},
  publisher={IEEE}
}

@article{soomro2012ucf101,
  title={UCF101: A dataset of 101 human actions classes from videos in the wild},
  author={Soomro, K},
  journal={arXiv preprint arXiv:1212.0402},
  year={2012}
}

@inproceedings{rasheed2023ivlp,
  title={Fine-tuned clip models are efficient video learners},
  author={Rasheed, Hanoona and Khattak, Muhammad Uzair and Maaz, Muhammad and Khan, Salman and Khan, Fahad Shahbaz},
  booktitle={Proceedings of the IEEE/CVF Conference on Computer Vision and Pattern Recognition},
  pages={6545--6554},
  year={2023}
}

@article{khattak2024protext,
  title={Learning to Prompt with Text Only Supervision for Vision-Language Models},
  author={Khattak, Muhammad Uzair and Naeem, Muhammad Ferjad and Naseer, Muzammal and Van Gool, Luc and Tombari, Federico},
  journal={arXiv preprint arXiv:2401.02418},
  year={2024}
}

@article{xu2024provp,
  title={Progressive visual prompt learning with contrastive feature re-formation},
  author={Xu, Chen and Zhu, Yuhan and Shen, Haocheng and Chen, Boheng and Liao, Yixuan and Chen, Xiaoxin and Wang, Limin},
  journal={International Journal of Computer Vision},
  pages={1--16},
  year={2024},
  publisher={Springer}
}

@inproceedings{guo2025mmrl,
  title={Mmrl: Multi-modal representation learning for vision-language models},
  author={Guo, Yuncheng and Gu, Xiaodong},
  booktitle={Proceedings of the Computer Vision and Pattern Recognition Conference},
  pages={25015--25025},
  year={2025}
}

@inproceedings{li2025dpc,
  title={Dpc: Dual-prompt collaboration for tuning vision-language models},
  author={Li, Haoyang and Wang, Liang and Wang, Chao and Jiang, Jing and Peng, Yan and Long, Guodong},
  booktitle={Proceedings of the Computer Vision and Pattern Recognition Conference},
  pages={25623--25632},
  year={2025}
}

@INPROCEEDINGS{li2025mao,
  author={Li, Haoyang and Zhou, Siyu and Wang, Liang and Long, Guodong},
  booktitle={2025 IEEE International Conference on Multimedia and Expo (ICME)}, 
  title={MAO: Efficient Model-Agnostic Optimization of Prompt Tuning for Vision-Language Models}, 
  year={2025},
  pages={1-6}
}

@article{li2025augpt,
  title={Raw Data Matters: Enhancing Prompt Tuning by Internal Augmentation on Vision-Language Models},
  author={Li, Haoyang and Wang, Liang and Wang, Chao and Zhou, Siyu and Jiang, Jing and Peng, Yan and Long, Guodong},
  journal={arXiv preprint arXiv:2508.02671},
  year={2025}
}

@inproceedings{li2025atprompt,
  title={Advancing textual prompt learning with anchored attributes},
  author={Li, Zheng and Song, Yibing and Cheng, Ming-Ming and Li, Xiang and Yang, Jian},
  booktitle={Proceedings of the IEEE/CVF International Conference on Computer Vision},
  pages={3618--3627},
  year={2025}
}

@article{zhang2025dapt,
  title={Decouple before align: Visual disentanglement enhances prompt tuning},
  author={Zhang, Fei and Zhou, Tianfei and Yao, Jiangchao and Zhang, Ya and Tsang, Ivor W and Wang, Yanfeng},
  journal={IEEE Transactions on Pattern Analysis and Machine Intelligence},
  year={2025},
  publisher={IEEE}
}

@inproceedings{selvaraju2017gradcam,
  title={Grad-cam: Visual explanations from deep networks via gradient-based localization},
  author={Selvaraju, Ramprasaath R and Cogswell, Michael and Das, Abhishek and Vedantam, Ramakrishna and Parikh, Devi and Batra, Dhruv},
  booktitle={Proceedings of the IEEE international conference on computer vision},
  pages={618--626},
  year={2017}
}

@article{zou2023seem,
  title={Segment everything everywhere all at once},
  author={Zou, Xueyan and Yang, Jianwei and Zhang, Hao and Li, Feng and Li, Linjie and Wang, Jianfeng and Wang, Lijuan and Gao, Jianfeng and Lee, Yong Jae},
  journal={Advances in neural information processing systems},
  volume={36},
  pages={19769--19782},
  year={2023}
}

@article{yang2023fgvp,
  title={Fine-grained visual prompting},
  author={Yang, Lingfeng and Wang, Yueze and Li, Xiang and Wang, Xinlong and Yang, Jian},
  journal={Advances in Neural Information Processing Systems},
  volume={36},
  pages={24993--25006},
  year={2023}
}

@inproceedings{shtedritski2023redcircle,
  title={What does clip know about a red circle? visual prompt engineering for vlms},
  author={Shtedritski, Aleksandar and Rupprecht, Christian and Vedaldi, Andrea},
  booktitle={Proceedings of the IEEE/CVF International Conference on Computer Vision},
  pages={11987--11997},
  year={2023}
}

@inproceedings{sun2024AlphaCLIP,
  title={Alpha-clip: A clip model focusing on wherever you want},
  author={Sun, Zeyi and Fang, Ye and Wu, Tong and Zhang, Pan and Zang, Yuhang and Kong, Shu and Xiong, Yuanjun and Lin, Dahua and Wang, Jiaqi},
  booktitle={Proceedings of the IEEE/CVF conference on computer vision and pattern recognition},
  pages={13019--13029},
  year={2024}
}

@article{xing2024peft2,
  title={A survey of efficient fine-tuning methods for vision-language models—prompt and adapter},
  author={Xing, Jialu and Liu, Jianping and Wang, Jian and Sun, Lulu and Chen, Xi and Gu, Xunxun and Wang, Yingfei},
  journal={Computers \& Graphics},
  volume={119},
  pages={103885},
  year={2024},
  publisher={Elsevier}
}

@article{an2023backgroundCLIP,
  title={More context, less distraction: Visual classification by inferring and conditioning on contextual attributes},
  author={An, Bang and Zhu, Sicheng and Panaitescu-Liess, Michael-Andrei and Mummadi, Chaithanya Kumar and Huang, Furong},
  journal={arXiv preprint arXiv:2308.01313},
  pages={7368--7377},
  year={2023}
}

@inproceedings{houlsby2019Adapter,
  title={Parameter-efficient transfer learning for NLP},
  author={Houlsby, Neil and Giurgiu, Andrei and Jastrzebski, Stanislaw and Morrone, Bruna and De Laroussilhe, Quentin and Gesmundo, Andrea and Attariyan, Mona and Gelly, Sylvain},
  booktitle={International conference on machine learning},
  pages={2790--2799},
  year={2019},
  organization={PMLR}
}

@inproceedings{zhuang2024falip,
  title={Falip: Visual prompt as foveal attention boosts clip zero-shot performance},
  author={Zhuang, Jiedong and Hu, Jiaqi and Mu, Lianrui and Hu, Rui and Liang, Xiaoyu and Ye, Jiangnan and Hu, Haoji},
  booktitle={European Conference on Computer Vision},
  pages={236--253},
  year={2024},
  organization={Springer}
}

@inproceedings{lu2022proda,
  title={Prompt distribution learning},
  author={Lu, Yuning and Liu, Jianzhuang and Zhang, Yonggang and Liu, Yajing and Tian, Xinmei},
  booktitle={Proceedings of the IEEE/CVF conference on computer vision and pattern recognition},
  pages={5206--5215},
  year={2022}
}

@inproceedings{kan2023kapt,
  title={Knowledge-aware prompt tuning for generalizable vision-language models},
  author={Kan, Baoshuo and Wang, Teng and Lu, Wenpeng and Zhen, Xiantong and Guan, Weili and Zheng, Feng},
  booktitle={Proceedings of the IEEE/CVF International Conference on Computer Vision},
  pages={15670--15680},
  year={2023}
}

@article{li2026hardnegative,
title = {Negative-Sampling Prompt Learning for Hard Negative Sample Discrimination},
journal = {Knowledge-Based Systems},
pages = {115603},
year = {2026},
issn = {0950-7051},
author = {Li, Haoyang and Wang, Liang and Peng, Yan and Wang, Chao}
}
